\documentclass[10pt,twocolumn,letterpaper]{article}

\usepackage{iccv}
\usepackage{times}
\usepackage{epsfig}
\usepackage{graphicx}
\usepackage{amsmath}
\usepackage{amssymb}
\usepackage{times}
\usepackage{multirow}
\usepackage{subfig}
\usepackage{amsthm}
\usepackage{bbm}
\usepackage[ruled,vlined]{algorithm2e}
\usepackage{makecell}
\usepackage{rotating}
\usepackage{booktabs}
\usepackage{float}
\usepackage{color}
\usepackage{bm}
\usepackage{subfloat}
\usepackage{threeparttable}

\newcommand{\vect}[1]{\mathbf{#1}}
\newcommand{\matr}[1]{\mathbf{#1}}

\newcommand{\set}[1]{\mathcal{#1}}

\usepackage[pagebackref=true,breaklinks=true,letterpaper=true,colorlinks,bookmarks=false]{hyperref}

\iccvfinalcopy 

\def\iccvPaperID{1362} 

\ificcvfinal\fi
\begin{document}

\title{Learning for Multi-Type Subspace Clustering}

\author{Xun Xu\\
National University of Singapore\\
{\tt\small alex.xun.xu@gmail.com}
\and
Loong-Fah Cheong\\
National University of Singapore\\
{\tt\small eleclf@nus.edu.sg}
\and
Zhuwen Li\\
Pony AI\\
{\tt\small lzhuwen@gmail.com}
}

\maketitle

\begin{abstract}
    Subspace clustering has been extensively studied from the hypothesis-and-test, algebraic, and spectral clustering-based perspectives. Most assume that only a single type/class of subspace is present. Generalizations to multiple types are non-trivial, plagued by challenges such as choice of types and numbers of models, sampling imbalance and parameter tuning. In this work, we formulate the multi-type subspace clustering problem as one of learning non-linear subspace filters via deep multi-layer perceptrons (mlps). The response to the learnt subspace filters serve as the feature embedding that is clustering-friendly, i.e., points of the same clusters will be embedded closer together through the network. For inference, we apply K-means to the network output to cluster the data. Experiments are carried out on both synthetic and real world multi-type fitting problems, producing state-of-the-art results\footnote{Tensorflow implementations and corresponding data will be released on GitHub.}.  
\end{abstract}

\section{Introduction}

Subspace clustering aims to cluster data points into separate subspaces, with the dimension of the subspaces typically much smaller than the ambient space. 
Examples include vanishing point detection \cite{Rother2002}, rigid motion segmentation \cite{sugaya2004geometric,torr1998geometric,xu2018motion} and face clustering \cite{ho2003} 
 To make the problem tractable, traditional subspace clustering approaches tend to make various assumptions, such as data lying on a linear manifold, independence between subspaces, data drawn from a single type of subspace, known number of models, etc.. 

\begin{figure}[!t]
\centering
\includegraphics[width=1.05\linewidth]{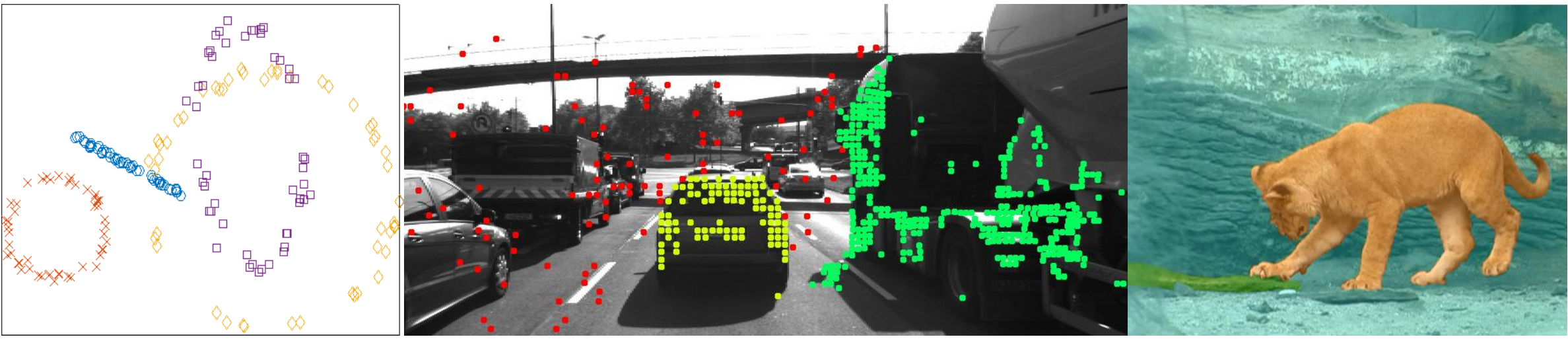}
\vspace{-0.5cm}
\caption{\small{Multi-type subspace clustering examples.}}\label{fig:MultiTypeIllust}
\vspace{-0.5cm}
\end{figure}

Despite the considerable amount of effort, there are still major lacunae in this research. Firstly, many real world problems consist of data drawn from a union of multiple types of subspaces.
We term this problem multi-type subspace clustering. Fig.~\ref{fig:MultiTypeIllust} shows some examples: a toy example of line, circle and ellipses co-existing together, and two real-world motion segmentation scenarios. In the latter two scenarios, the appropriate model to fit the foreground object motions can waver between affine motions, homography, fundamental matrix \cite{xu2018motion}, and even non-rigid motion, with no clear dividing boundary between them. 
With few exceptions \cite{Barath2018,sugaya2004geometric,torr1998geometric}, none of the existing works have considered this realistic scenario. Even if one attempts to fit multiple types of model sequentially like in \cite{sugaya2004geometric}, it is non-trivial to decide the type when the dichotomy of the models is unclear in the first place, e.g. when is the rotation dominant enough so that homography becomes a better model than fundamental matrix? For non-rigid motions, an analytic subspace model can be hard to define, thus neither the hypothesize-and-test nor the algebraic approach could be easily applied.

Secondly, for problems where there are a significant number of models, the traditional hypothesis-and-test approach is often overwhelmed by sampling imbalance, i.e., points from the same subspace represent only a minority, rendering the probability of hitting upon the correct hypothesis very small. This problem becomes severe when a large number of data samples are required for hypothesizing a model (e.g., eight points are needed for a linear estimation of the fundamental matrix and 5 points for fitting an ellipse). Moreover, for optimal performance, there is inevitably a lot of manipulation of parameters needed, among which the most sensitive include those for deciding what constitutes an inlier for a model \cite{Magri2014, Magri2015}, for sparsifying the affinity matrices \cite{Lai2017,xu2018motion}, and for selecting the model type \cite{torr1998geometric}. Often, dataset-specific tuning is required, with very little theory to guide the tuning. 

Another open challenge in subspace clustering is to automatically determine the number of models, also referred to as model selection in the literature \cite{Tibshirani2001,Chin2009,Li_2013_ICCV,Lai2017}. Traditional methods are based upon the statistical analysis of the residual of the clustering \cite{Tibshirani2001,rousseeuw1987silhouettes}. 
Other methods approach the problem using various heuristics including analyzing eigen values \cite{zelnik2005self,von2007tutorial}, over-segment and merge \cite{Li_2013_ICCV,Lai2017}, soft thresholding \cite{liu2013robust} or adding penalty terms \cite{Li2014}. 
Most of the above works require extensive parameter tuning and have never been tested on data drawn from mixed-type of models. 
Lastly, hypothesis-and-test methods have to go through expensive sampling step, whereas analytic approaches have to contend with solving complex optimization problems. Thus, both approaches suffer from slow inference (as evidenced by our experimental comparisons), which is a serious qualification for real-time applications.

With the above considerations, we propose the SubspaceNet, a deep network that learns appropriate feature embeddings from input feature points without having to manually design similarity metric nor to know the subspace model a priori. The learnt feature representation allows clusters to be readily identified using off-the-shelf methods, even when the underlying data are drawn from a union of mixed types of models, with the dividing boundary between these multiple types of subspaces being unclear (e.g. the transitions from a circle to an ellipse), or the underlying subspace is not analytically expressible (e.g. non-rigid motion). Our network consists mainly of stacked multi-layer perceptions (mlps). 
Each of the mlps has output in the form of $y=\matr{w}^\top x+b$, which describes a linear subspace. For each layer of mlp($m$,$n$) ($m$ and $n$ indicate the number of input and output neurons respectively), we have up to $n$ different subspaces and they could be stacked together to define convex polytopes delimited by multiple linear cuts in the original space. More importantly, by coupling mlps with non-linear activations functions and stacking the resultant nonlinear features into a hierarchy, we can approximate very complex non-linear subspaces in the ambient space. 
At each layer of mlp, feature points are represented as responses (distances) to the subspaces. This is analogous to the concept of Ordered Residual Kernel (ORK) in \cite{Chin2009}: feature points of the same model display similar responses to the set of subspaces hypothesized and these responses can be regarded as a new form of feature representation. 
Here, given labelled data (inlier points for each model and outliers), the network learns the appropriate subspace filters (mlps) that produce the feature embeddings (responses to mlps) amenable for grouping into the respective, possibly mixed models. The preference for the various mixed types of models is also decided by the network in a data-driven manner without having to tune a lot of system parameters. 

We summarize our contributions as follows. (i) First, we address multi-type subspace clustering, i.e. data drawn from mixed types of (possibly non-analytic) models. (ii) Our solution naturally affords the ability to handle  model selection and sampling imbalance. 
(iii) We propose a subspace clustering network (SubspaceNet) by stacking multi-layer perceptrons and achieved state-of-the-art performance on three datasets. The SubspaceNet is more effective than alternative networks designed for sparse set of feature points. (iv) We proposed a more effective metric learning loss optimizing the distribution of learnt feature embedding. 
(v) Our SubspaceNet is highly efficient at the inference stage and outperforms existing non-deep approaches by a large margin.

\section{Related Work}

\noindent\textbf{Subspace Fitting}:
Early approaches address this in a sequential RANSAC fashion \cite{torr1998geometric,vincent2001detecting,Yasushi2004} by iteratively fitting and removing inliers. 
The J-Linkage \cite{toldo2008robust} and T-Linkage \cite{Magri2014} simultaneously consider the interactions between all points and hypotheses. The final partition is achieved by clustering. The above greedy algorithms often do not perform well under high noise level. Global algorithms have also been proposed to minimize an energy with various regularization terms, including spatial regularization (PEaRL) \cite{Isack2012} and label count penalty \cite{li2007two}. 
To eschew the problem of having to set thresholds, the ORK approach \cite{Chin2009,chin2010accelerated} ranked the hypothesis according to data preference rather than absolute residuals. 
Analytic approaches are characterized by elegant mathematical formulation, including those based on the sparsity \cite{Elhamifar2013} and low-rank \cite{liu2013robust} assumptions and their variants. Many of the preceding works adopt spectral clustering for final grouping and assume known number of models, but only a few considered the model selection problem, e.g., \cite{Alzate2010,Li_2013_ICCV,liu2013robust,Soltanolkotabi2011}. Even fewer works \cite{Barath2018,goh2007segmenting,sugaya2004geometric,torr1998geometric} considered the problem of fitting multiple model of various types, and in these few works, the types are assumed to be known a priori and well-defined which is often not realistic. 

\noindent\textbf{Deep Learning for Geometric Modeling Problems}:
Using deep learning to solve geometric model fitting has received growing considerations. The dense approaches use raw image to model the transformation between image pairs as homography \cite{detone2016deep} or non-rigid transformation \cite{rocco2017convolutional}. \cite{melekhov2017relative} proposed to estimate the camera pose directly from image sequences. 
DSAC\cite{brachmann2017dsac} learns to extract from sparse feature correspondences a geometric model in a manner akin to RANSAC. The ability to learn representations from sparse points was also developed recently\cite{Qi2017,QiLSG_NIPS2017}. This ability was exploited by \cite{yi2018learning} to fit essential matrix from noisy correspondences. Despite the promising results, none of the existing works have considered generic model fitting and, more importantly, fitting data drawn from multiple models and even multiple types. In our work, we formulate the generic multi-type fitting problem as one of learning good representations for clustering.

\noindent\textbf{Deep Learning for Clustering}:
Unsupervised approaches tackle the problem by finding a latent embedding that minimizes the reconstruction loss of an autoencoder \cite{huang2007unsupervised,vincent2008extracting,lee2009convolutional}.  They are further combined with various losses for clustering objectives \cite{tian2014learning,yang2016joint,ji2017deep,xie2016unsupervised}. Among these, the k-means loss was proposed by \cite{yang2016joint} optimizing the points-to-center distance. 
The subspace self-expressiveness objective was considered for discovering linear subspaces in the latent space \cite{ji2017deep}.
In our tasks, there is a multiplicity of geometric models that are equally valid and subtly differentiated. For instance, given images of a rigidly moving cube, are we supposed to group the trajectory features by a single fundamental matrix or by multiple homographies? 
It would be difficult for the unsupervised networks to know the preference without any form of supervision. 

In supervised approach, labelled data are used to learn feature embedding amenable for clustering \cite{Xing2003}. 
With the advent of deep neural network, works in metric learning focus on designing losses amenable to clustering labelled data\cite{Chopra2005, Schroff2015, Sohn2016, hershey2016deep, song2017deep}. 
Among these, \cite{hershey2016deep} minimizes the L2 distance between the predicted and ground-truth affinities and provides a competitive baseline. To further take into account the global distribution of the data points, we propose the clustering-specific loss MaxInterMinIntra, which optimizes the inter-cluster separation and intra-cluster variance and is proven to be more effective than existing alternatives.

\section{Methodology}


\begin{figure}[!ht]
\vspace{-0.2cm}
\begin{center}
\includegraphics[width=1.03\linewidth]{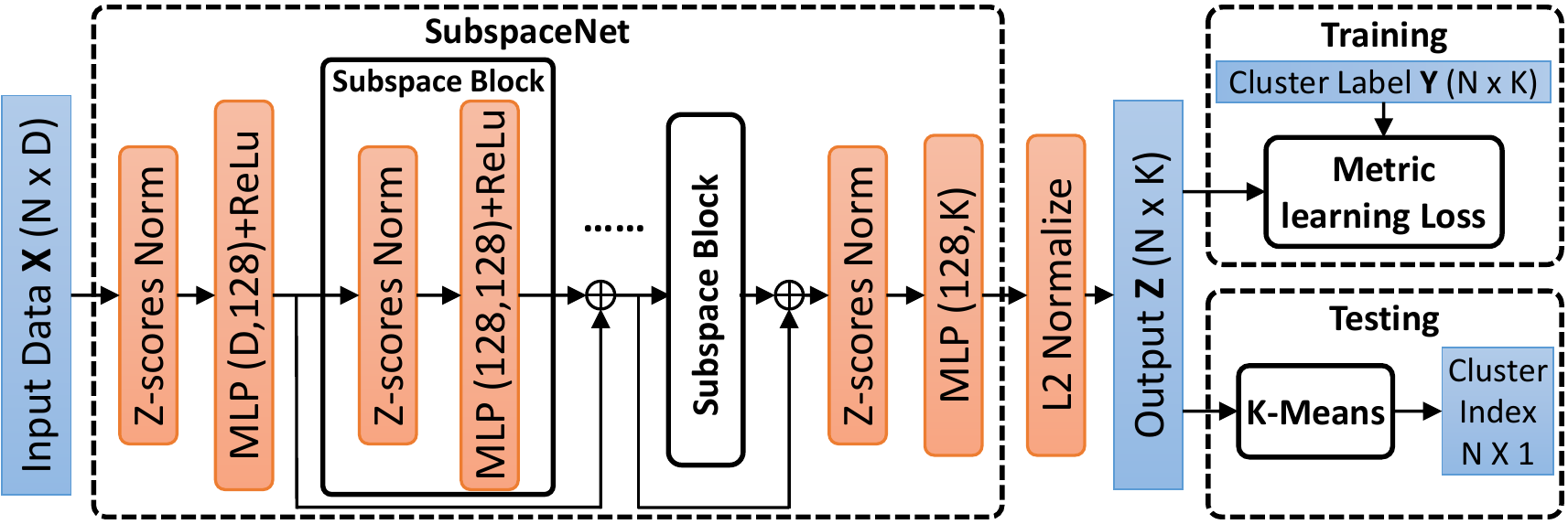}
\caption{\small{Our Subspace clustering network. The metric learning loss is defined to learn good feature representation.} }\label{fig:Network}
\end{center}\vspace{-0.7cm}
\end{figure}

\subsection{ Network Architecture}

We denote the input sparse data with $N$ points as $\matr{X}=\{\vect{x}_{i}\}_{i=1\cdots N}\in\set{R}^{D\times N}$ where each individual point is $\vect{x}_{i}\in\set{R}^D$. The input sparse data could be geometric shapes, feature correspondences in two frames or feature trajectories in multiple frames. We further denote the one-hot key encoded labels accompanying the input data as $\matr{Y}=\{\vect{y}_{i}\}\in\{0,1\}^{K\times N}$ where $\matr{y}_{i}\in \{0,1\}^{K}$ and $K$ is the number of clusters or partitions of the input data. 

Our subspace network consists mainly of stacked multi-layer perceptions (mlps) as shown in Fig.~\ref{fig:Network}. It resembles the correspondence network\cite{yi2018learning} in that both exploit the power of mlps. We have noted that each layer of mlp works as multiple linear subspaces and the response to each layer of mlp serves as the new feature representation of the input feature points. Since the mlps are not scale invariant, a normalization layer is thus necessary before each mlp layer to center all feature points at origin with unit variance. The is realized by a standard z-score normalization on each input dimension, denoted as \textit{Z-score Norm} layer in Fig.~\ref{fig:Network}. We note that this step resembles the context norm (CN) proposed in \cite{yi2018learning}. However, the role of CN was ascribed to capturing the relation between feature points by \cite{yi2018learning} whereas here, we believe the role of \textit{Z-score Norm} is more specifically that of ensuring uniform scale. We adopt the same ResNet\cite{he2016deep} structure with CorresNet for training deeper network and the depth, number of \textit{Subspace Blocks} is fixed at 50 for all experiments. For the output layer, we do not apply any activation but instead conduct L2 normalization on each sample. The output embedding is denoted as $\matr{Z}=\{f(\matr{X};\Theta)\}\in\set{R}^{K\times N}$. To make the output $\matr{Z}$ clustering-friendly, we apply a differentiable, clustering-specific loss function $\mathcal{L}(\matr{Z},\matr{Y})$,  
measuring the match of the output feature representation with the ground-truth labels.  The problem now becomes that of learning a CorresNet backbone $f(\matr{X};\Theta)$ that minimizes the loss $\mathcal{L}(\matr{Z},\vect{Y};\Theta)$.


%

\subsection{Clustering Loss}
We expect our clustering loss function to have the following characteristics. First, it should be invariant to permutation of models, e.g. the order of these models are exchangeable. Second the loss must be adaptable to varying number of groups. Lastly, the loss should enable good separation of data points into clusters. We consider the following loss functions. 

\noindent\textbf{L2Regression Loss}:
Given the ground-truth labels $\vect{Y}$ and the output embeddings $\matr{Z}=f(\vect{X};\Theta)$, the ideal and reconstructed affinity matrices are respectively,
\begin{equation}
\vspace{-0.2cm}
\begin{split}
&\matr{K} = \vect{Y}^\top\vect{Y}, \quad \matr{\hat{K}} = \matr{Z}^\top\matr{Z}
\end{split}
\vspace{-0.2cm}
\end{equation}

The training objective is to minimize the difference between $\matr{K}$ and $\matr{\hat{K}}$ measured by element-wise L2 distance \cite{hershey2016deep}.
\vspace{-0.2cm}
\begin{equation}
\begin{split}
L(\Theta) &= ||\matr{K}-\matr{\hat{K}}||_F^2 \\
&= ||\matr{Y}^\top\matr{Y} - \vect{Z}^\top\vect{Z}||_F^2\\
&= ||f(\matr{X};\Theta)^\top f(\matr{X};\Theta)||_F^2 - 2||f(\matr{X};\Theta)\vect{Y}^\top||_F^2
\end{split}
\end{equation}

The above L2 Regression loss is obviously differentiable w.r.t. $f(\matr{X};\Theta)$. Since the output embedding $\matr{Z}$ is L2-normalized, the inner product between two point representations is $\vect{z}_i^\top\vect{z}_j\in[-1,1]$. 

\noindent\textbf{Cross-Entropy Loss}: As alternative to the L2 distance, one could measure the discrepancy between $\matr{K}$ and $\hat{\matr{K}}$ as KL-Divergence. Since $D_{kl}(\matr{K}||S(\matr{\hat{K}}))=H(\matr{K},S(\matr{\hat{K}})) - H(\matr{K})$, where $H(\cdot)$ is the entropy function and $S(\cdot)$ is the sigmoid function, with fixed $\matr{K}$, we simply need to minimize the cross-entropy $H(\matr{K},S(\matr{\hat{K}}))$ which yields the following element-wise cross-entropy loss,

\begin{equation}
\vspace{-0.1cm}
\begin{split}
L(\Theta)&= \sum_{i,j} H\left(\vect{y}_{i}^\top\vect{y}_{j},S\left(\vect{z}_{i}^\top\vect{z}_{j}\right)\right) \\
&= \sum_{i,j} H(\vect{y}_{i}^\top\vect{y}_{j},S(f(\vect{x}_i;\Theta)^\top f(\vect{x}_i;\Theta)))
\end{split}
\end{equation}

\vspace{-0.1cm}
 The cross-entropy loss is more likely to push points $i$ and $j$ of the same cluster together faster than L2Regression, i.e. inner product $\vect{z}_i^\top\vect{z}_j \rightarrow 1$ and those of different clusters apart, i.e. inner product $\vect{z}_i^\top\vect{z}_j \rightarrow -1$.

\noindent\textbf{MaxInterMinIntra Loss}:
Both the above losses consider the pairwise relation between points; the overall point distribution in the output embedding is not explicitly considered. We now propose a new loss which takes a more global view of the point distribution rather than just the pairwise relations. Specifically, we are inspired by the classical Fisher LDA \cite{fisher1936use}. LDA discovers a linear mapping ${z}=\vect{w}^\top\vect{x}$ that maximizes the distance between class centers/means $\mu_i=1/N \sum_j {z}_j$ and minimizes the scatter/variance within each class $s_i = \sum_j(z_j-\mu_i)^2$. Formally, the objective for a two-class problem is written as,
\begin{equation}
J(\vect{w})=  \frac{|\mu_1-\mu_2|^2}{s_1^2+s_2^2}
\end{equation}
which is to be maximized over $\vect{w}$. 
For linearly non-separable problem, one has to design kernel function to map the input features before applying the LDA objective. Equipped now with more powerful nonlinear mapping networks, we adapt the LDA objective---for the multi-class scenarios---to perform these mappings automatically as below, 
\begin{equation}
J(\Theta)=\frac{\min\limits_{m,n\in\{1\cdots K\}, m\neq n}||\bm{\mu}_m - \bm{\mu}_n||^2_2}{\max\limits_{l\in\{1\cdots K\}} s_l}
\end{equation}
where $\bm{\mu}_m=\frac{1}{|\mathcal{C}_m|}\sum_{i\in\mathcal{C}_m} \vect{z}_i$, $s_l = \sum_{i\in\mathcal{C}_l}||\vect{z}_i-\bm{\mu}_l||^2_2$ and $\mathcal{C}_l$ indicating the set of points belonging to cluster $l$.  We use the extrema of the inter-cluster distances and intra-cluster scatters (see Fig.~\ref{fig:MaxInterMinIntra}) so that the worst case is explicitly optimized. Hence, we term the loss as MaxInterMinIntra (MIMI). By applying log operation on the objective,
we arrive at the following loss function to be minimized: 

\begin{figure}
\includegraphics[width=1\linewidth]{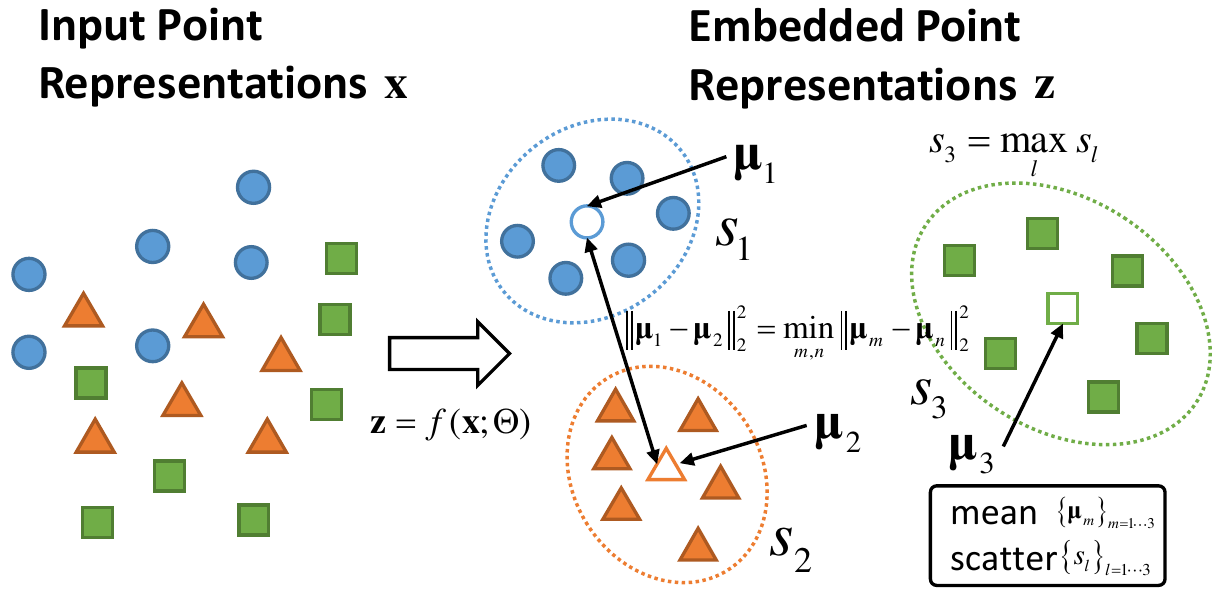}
\caption{\small{Illustration of MaxInterMinIntra loss for point representation metric learning. The objective considers the minimal distance $\min_{m,n}||\bm{\mu}_m-\bm{\mu}_n||_2^2$ between clusters and maximal scatter  $\max_l s_l$ within clusters.}}\label{fig:MaxInterMinIntra}
\vspace{-0.5cm}
\end{figure}

\begin{equation}\label{eq:MaxInterMinIntra}
L(\Theta) = -\log \min\limits_{m,n} ||\bm{\mu}_m-\bm{\mu}_n||_2^2 + \log\max_{l} s_l
\end{equation}
\noindent
One can easily verify that the MaxInterMinIntra loss is differentiable w.r.t. $\vect{z}_i$. We provide the gradient in the supplementary material. 

\noindent\textbf{Optimization}:
The Adam optimizer \cite{kingma2014adam} is used to minimize the loss $L(\Theta)$. The learning rate is fixed at $1e-3$ and mini-batch at one frame pair or sequence. 
 For all tasks, we train the network for 300 epochs. 


\subsection{Inference}
During testing, we apply standard K-means to the output embeddings $\{\vect{z}_j\}_{j=1\cdots N_{te}}$. 
This step is applicable to both multi-model and multi-type clustering problems, as we do not need to specify explicitly the type of model to fit. If there is a need to estimate the number of models $K$, we examine the K-means residuals defined by, 
\begin{equation}
r(K) = \sum_{m=1\cdots K}\sum_{i\in\mathcal{C}_m} ||\vect{z}_i - \bm{\mu}_m||_2^2
\end{equation}

Good estimate of $K$ often yields low $r(K)$ and further increasing $K$ does not significantly reduce $r(K)$. Thus we find the $K$ at the `elbow' position. We adopt two off-the-shell approaches for this purpose:  second order difference (SOD)\cite{Zhang2012} and silhouette analysis \cite{rousseeuw1987silhouettes}. Both are parameter-free.

\section{Experiment}

We demonstrate the performance of our network on both synthetic and real world data, with extensive comparisons with traditional geometric model fitting algorithms. 

\subsection{Datasets}


\noindent\textbf{Synthesized Lines, Circles and Ellipses (LCE)}: Fitting ellipses has been a fundamental problem in computer vision \cite{fitzgibbon1999direct}. We synthesize for each sample four different types of conic curves in a 2D space, specifically, one straight line, two ellipses and one circle. We randomly generate 8,000 training samples, 200 validation samples and 200 testing samples. Each point is perturbed by adding a gaussian noise with $\sigma=0.05$.

\noindent\textbf{KT3DMoSeg} \cite{xu2018motion}: 
This benchmark consists of 22 sequences from the KITTI dataset\cite{Geiger2013IJRR}.
Each sequence contains two to five rigid motions. As analyzed by \cite{xu2018motion}, the geometric model for each individual motion can range from an affine transformation, a homography, to a fundamental matrix, with no clear dividing line between them. We evaluate this benchmark to demonstrate our network's ability to tackle multi-type clustering. For fair comparison, we only crop the first 5 frames of each sequence for evaluation, so that the broken trajectory does not give undue advantage to certain methods.

\noindent\textbf{FBMS59 \cite{ochs2014segmentation}}: This dataset was proposed for analyzing video object segmentation based on point trajectories, with 59 sequences in total, of which 29 are for training and 30 for testing. It covers a wide variety of scenes and the ground-truth is defined over semantic objects with dense mask. Most of the moving objects involve moderate non-rigidity, for which analytic geometric models are hard to define. We evaluate the first-10-frame setting as reported in \cite{ochs2014segmentation} for fair comparison. The ground-truth for training is constructed by assigning the trajectories to the nearest label mask and the evaluation metric is the standard F-measure \cite{ochs2014segmentation}.




\noindent\textbf{Adelaide RMF Dataset} \cite{wong2011dynamic}: We are concerned with the two-view motion segmentation task of this dataset. This task consists of 19 frame pairs each comprising of 2 to 5 independent motions. Though it is nominally a single-type multiple fundamental matrix fitting problem and has been treated as such by the community, we observe moderate degeneracies, i.e. near planar rigid objects, present in this dataset. Hence, we treat it as another multi-type (homography and fundamental matrix) clustering problem. 

\subsection{Multi-Type Curve Fitting}

There is no clear dividing boundary between lines, circles, and ellipses as they can be all explained by the general conic equation (with the special cases of lines and circles obtained by setting some coefficients to $0$):

\begin{equation}\label{eq:Conic}
Ax^2+Bxy+Cy^2+Dx+Ey+F=0
\end{equation}

There are two ways to adapt the traditional multi-model fitting methods for this multi-type setting. One approach formulates the problem as fitting multiple models parameterized by the same conic equation in Eq~(\ref{eq:Conic}), which is termed \textit{HighOrder} (H.O.) fitting.
Alternatively, one could sequentially fit three types of models, which is termed \textit{Sequential} (Seq.) fitting. For ellipse-specific fitting, the direct least square approach \cite{fitzgibbon1999direct} is adopted.
For our model, we evaluate the various metric learning losses introduced in Section~3.2 and present the results in Tab.~\ref{tab:SyntheticMultiType}. 
The results are reported with the optimal setting determined by the validation set. We evaluate the performance by two clustering metrics, Classification Error Rate (Error Rate), i.e. the best classification results subject to permutation of clustering labels, and Normalized Mutual Information (NMI). Comparisons are made with state-of-the-art multi-model fitting algorithms including T-linkage \cite{Magri2014}, RPA \cite{Magri2015} and RansaCov \cite{magri2016multiple}. We notice that T-linkage returns extremely over-segmented results in the sequential setting, e.g. more than 10 lines, making classification error evaluation intractable. 
For our model, we evaluate the three loss variants, the L2 Regression loss (L2), Cross Entropy loss (CE) and MaxInterMinIntra loss (MIMI).


\begin{table}[htbp]
  \centering
\caption{\small{Evaluations on synthetic multi-model and multi-type fitting dataset. $\uparrow$ and $\downarrow$ indicate the number is the higher or lower the better respectively. $-$ indicates evaluation intractable.}}\vspace{-0.2cm}
   \setlength\tabcolsep{2pt} 
  \resizebox{0.95\linewidth}{!}{
    \begin{tabular}{p{2.0em}ccccccccc}
    \toprule
    \multirow{2}[4]{*}{\textbf{Mdl.}} & \multicolumn{2}{c}{T-Linkage\cite{Magri2014}} & \multicolumn{2}{c}{RPA\cite{Magri2015}} & \multicolumn{2}{c}{RansaCov\cite{magri2016multiple}} & \multicolumn{3}{c}{SubspaceNet} \\
\cmidrule(lr){2-3}\cmidrule(lr){4-5}\cmidrule(lr){6-7} \cmidrule{8-10}   \multicolumn{1}{c}{} & \multicolumn{1}{c}{H.O.} & \multicolumn{1}{c}{Seq.} & \multicolumn{1}{c}{H.O.} & \multicolumn{1}{c}{Seq.} & \multicolumn{1}{c}{H.O.} & \multicolumn{1}{c}{Seq.} & \multicolumn{1}{c}{L2} & \multicolumn{1}{c}{CE} & \multicolumn{1}{c}{MIMI} \\
    \midrule
    \textbf{Err}$\downarrow$ & 52.14 & -  & 39.43 & 23.17 & 40.57 & 24.04 & 18.49 & 18.32 & \textbf{18.04} \\
    \textbf{NMI}$\uparrow$  & 0.340 & - & 0.464 & 0.667 & 0.394 & 0.604 & 0.713 & 0.720 & \textbf{0.727} \\
    \bottomrule
    \end{tabular}%
    }
  \label{tab:SyntheticMultiType}%
  \vspace{-0.3cm}
\end{table}%

\begin{figure}
\centering
\includegraphics[width=1\linewidth]{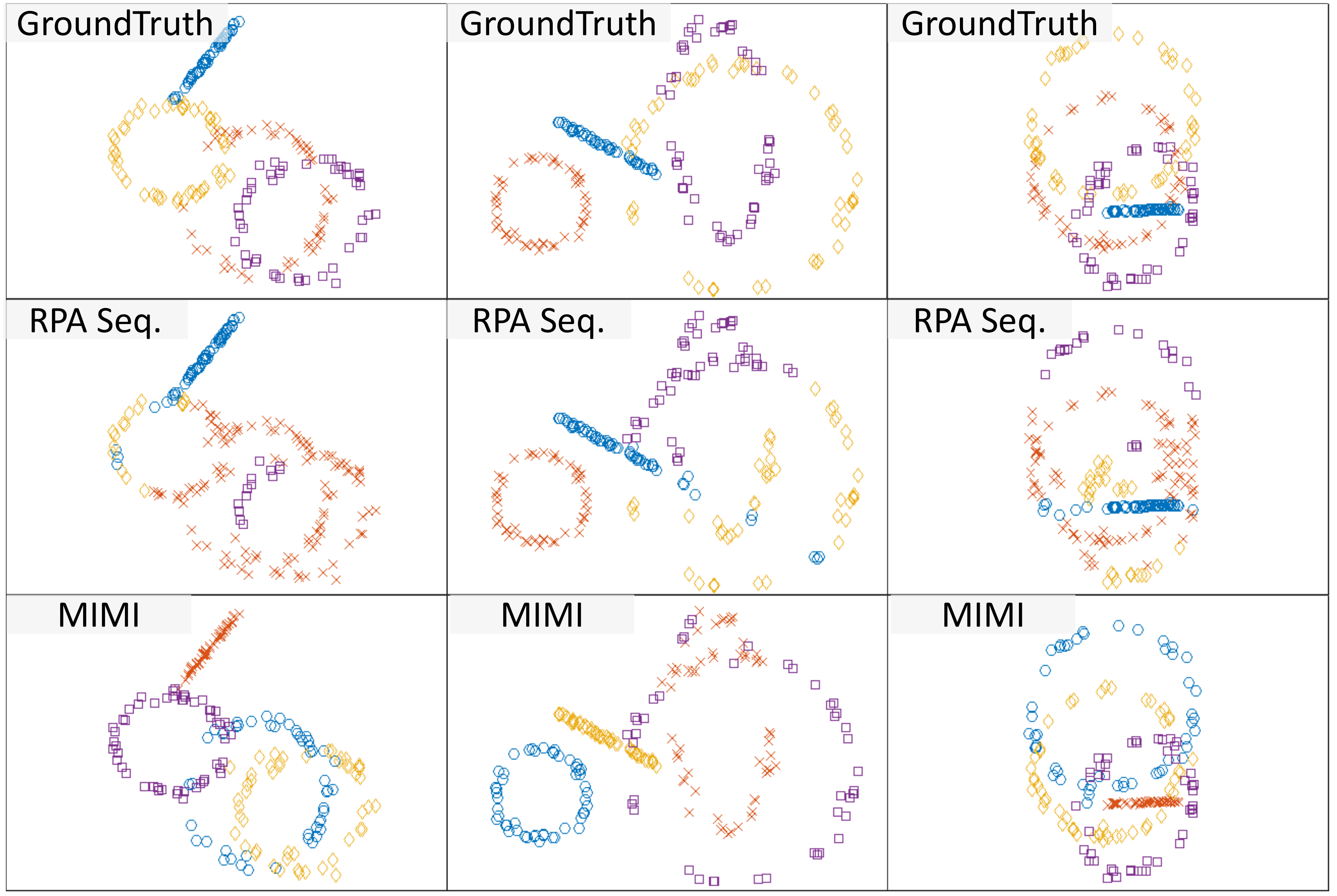}\vspace{-0.2cm}
\caption{\small{Examples of multi-type clustering on synthetic dataset. We only show the RPA results based on the \textit{Sequential} fitting approach.}}\label{fig:MultiTypeExamples}
\vspace{-0.5cm}
\end{figure}

We make the following observations about the results. First, all our metric learning variants outperform the \textit{HighOrder} and \textit{Sequential} multi-type fitting approaches. Second, the all-encompassing model used in the \textit{HighOrder} approach suffers from ill-conditioning when fitting simpler models. Thus, the performance is much inferior to that of \textit{Sequential} fitting. However, it is worth noting that  despite the \textit{Sequential} approach being given the strong a priori knowledge of both the model type and the number of model for each type, its performance is still significantly worse off than ours. 
For qualitative comparison, we visualize the ground-truth and segmentation results of each method in Fig.~\ref{fig:MultiTypeExamples}. Our clustering results on the bottom row show success in discovering all individual shapes with mistakes made only at the intersections of individual structures. The RPA failed to discover ellipses as sampling all 5 inliers amidst the large number of outliers and fitting an ellipse from even correct 5 support points with noise (noise in coordinate) are both very difficult, the latter demonstrated in \cite{fitzgibbon1999direct}. 

\begin{figure}[!htb]
\centering
\includegraphics[width=1.06\linewidth]{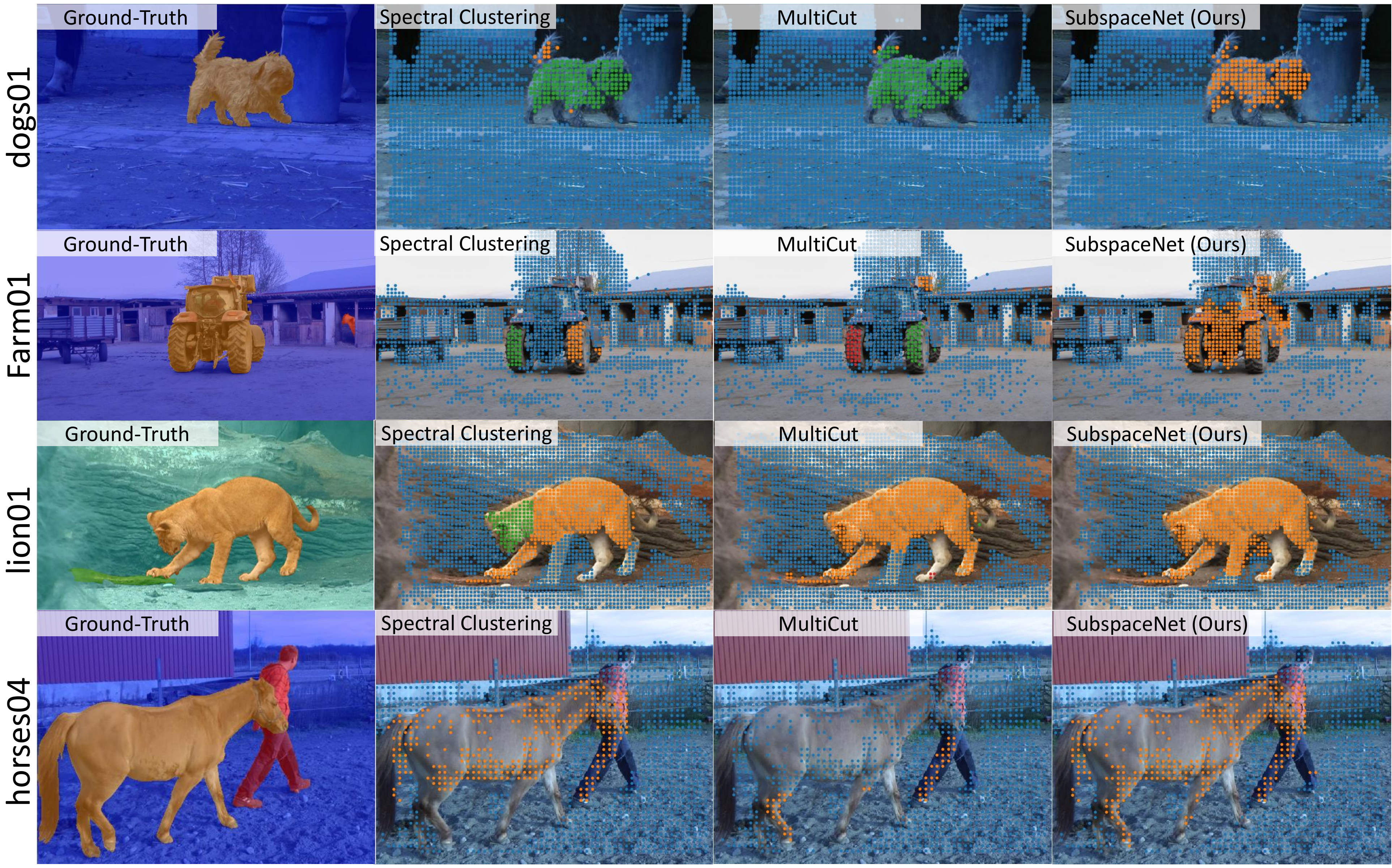}\vspace{-0.2cm}
\caption{\small{Qualitative comparisons on FBMS test set using the first 10 frames.}}\label{fig:FBMS_Qualitative}
\vspace{-0.5cm}
\end{figure}

\begin{table*}[htbp]
  \centering
  \caption{\small{Motion segmentation performance on KT3DMoSeg 5-frame task. Performances without (`Vanilla') and with augmentation (`Augment') are separated by a $/$. All error numbers are in $\%$ and inference time (Inf. Time) is in seconds.}}
  \vspace{-0.3cm}
     \footnotesize
   \setlength\tabcolsep{1pt} 
	  \resizebox{1.01\linewidth}{!}{
    \begin{tabular}{p{4em}ccccccccccccccc}
    \toprule
    \multicolumn{1}{c}{\multirow{3}[6]{*}{Model}} & \multicolumn{7}{c}{Non-Deep Approaches}               &       &       &       &       &       & \multicolumn{1}{l}{Deep Approaches} &       &  \\
\cmidrule(lr){2-8} \cmidrule(lr){9-16}    \multicolumn{1}{c}{} & \multicolumn{1}{c}{\multirow{2}[4]{*}{GPCA\cite{vidal2005generalized}}} & \multicolumn{1}{c}{\multirow{2}[4]{*}{ALC\cite{Rao2010}}} & \multicolumn{1}{c}{\multirow{2}[4]{*}{LSA\cite{Yan2006}}} & \multicolumn{1}{c}{\multirow{2}[4]{*}{LRR\cite{liu2013robust}}} & \multicolumn{1}{c}{\multirow{2}[4]{*}{MSMC\cite{Dragon2012}}} & \multicolumn{1}{c}{\multirow{2}[4]{*}{SSC\cite{Elhamifar2013}}} & \multicolumn{1}{c}{\multirow{2}[4]{*}{MVC\cite{xu2018motion}}} & \multicolumn{2}{c}{Unsupervised} & \multicolumn{6}{c}{Supervised (Different Losses)} \\
\cmidrule(lr){9-10}  \cmidrule(lr){11-16}   \multicolumn{1}{c}{} &       &       &       &       &       &       &       & \multicolumn{1}{c}{DSCN\cite{Ji2016}} & \multicolumn{1}{c}{DCN\cite{yang2017towards}} & \multicolumn{1}{c}{SHT\cite{schroff2015facenet}} & \multicolumn{1}{c}{LIFT\cite{oh2016deep}} & \multicolumn{1}{c}{NMI\cite{oh2017deep}} & \multicolumn{1}{c}{L2} & \multicolumn{1}{c}{CE} & \multicolumn{1}{c}{MIMI} \\
    \midrule
    Mean Err. & 36.46 & 15.17 & 36.34 & 22.00 & 32.74 & 26.62 & \textbf{10.99} & 28.14 & 48.45 & 10.93/6.88  & 28.06/21.25 & 16.91/10.92 & 10.95/6.89 & 11.12/7.81 & \textbf{10.62/5.83} \\
    Med. Err. & 33.93 & 16.42 & 40.31 & 18.16 & 36.48 & 29.14 & \textbf{6.57} & 30.00 & 48.16 & 9.11/5.02  & 28.34/22.69 & 11.65/8.54 & 7.84/3.83 & \textbf{7.04}/5.41 & 8.44/\textbf{3.58} \\
    Inf. Time & \textbf{1.51}  & 582.48 & 30.08 & 4.63  & 125.73 & 3254.99 & 143.52 & 1.85  & 1.85  & 1.85  & 1.85  & 1.85  & 1.85  & 1.85  & \textbf{1.85} \\
    \bottomrule
    \end{tabular}%
    }
  \label{tab:KT3DMoSeg}
  \vspace{-0.4cm}
\end{table*}%

\begin{figure*}[!htb]
\centering
\includegraphics[width=1\linewidth]{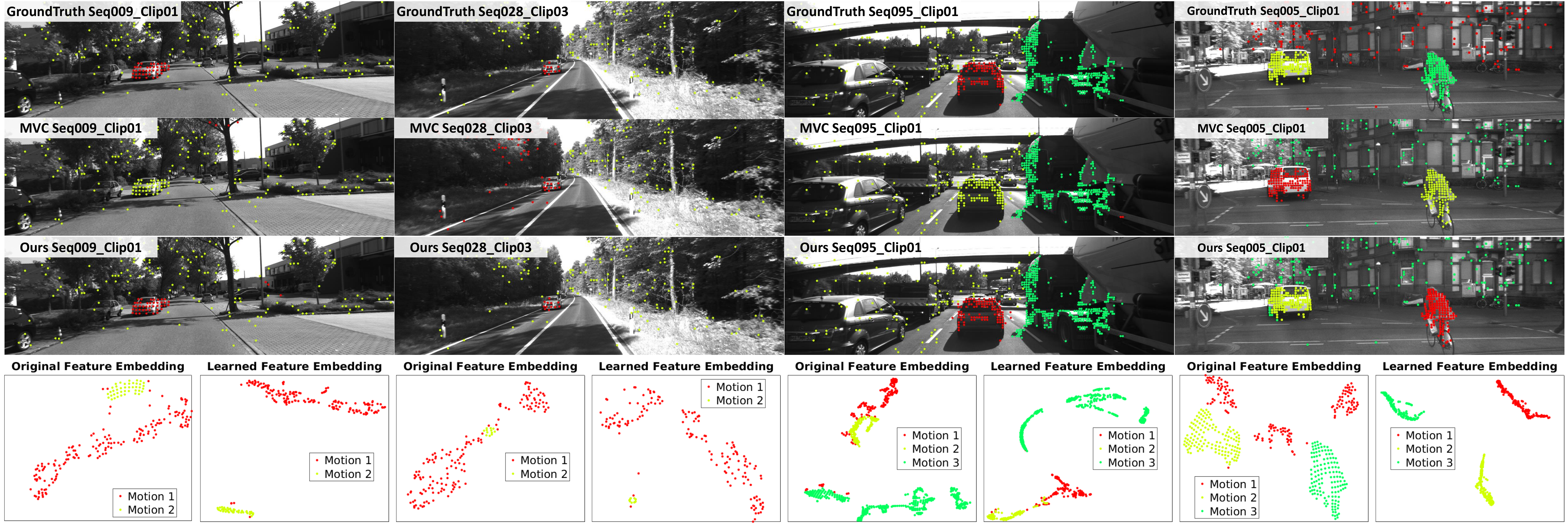}\vspace{-0.2cm}
\caption{\small{Qualitative comparison on 4 sequences from KT3DMoSeg. First row is the ground-truth. Second and third rows are the results of Multi-View Clustering \cite{xu2018motion} and our multi-type network respectively. The last row is the point feature embeddings before and after learning.}}\label{fig:KT3DMoSeg}\vspace{-0.5cm}
\end{figure*}

\subsection{Multi-Type Motion Segmentation}

Each sequence of the KT3DMoSeg benchmark\cite{xu2018motion} often consists of a background whose motion can be  explained  by a fundamental matrix while the models for the foreground motions can sometimes be ambiguous due to the limited spatial extent of the objects, thus giving rise to mixed types of models. 
For example, in Fig.~\ref{fig:KT3DMoSeg}, the vehicles in `Seq009\_Clip01' and `Seq028\_Clip03' can be roughly explained by an affine transformation or homography while the oil tanker in `Seq095\_Cip01' should be modeled by a fundamental matrix. When the background is dominated by a plane, for instance, the quasi-planar row of trees on the right side of the road in `Seq028\_Clip03', it is likely to lead to degeneracies in the fundamental matrix estimation. For this dataset, we apply leave-one-out cross-validation; we dubbed this  the `Vanilla' setting. Each sequence has between 10-20 frames, so we could further increase the training data by augmenting with all the remaining five-frame clips from each sequence, termed as the `Augment' setting. The testing clips (first five frames of each sequence) are kept the same for both settings. We compare with conventional non-deep subspace clustering approaches, GPCA\cite{vidal2005generalized}, LSA\cite{Yan2006}, ALC\cite{Rao2010}, LRR\cite{liu2013robust}, MSMC\cite{Dragon2012} and SSC\cite{Elhamifar2013} and the multi-view clustering (MVC) methods in \cite{xu2018motion}. For the unsupervised deep clustering approaches, we include the Deep Subspace Clustering Network (DSCN)\cite{Ji2016} and Simultaneous Deep Learning and Clustering (DCN)\cite{yang2017towards} for comparison. 
For the supervised setting, 
we compare with semi-hard triplet loss (SHT) \cite{schroff2015facenet}, lifted structured feature embedding (LIFT)\cite{oh2016deep} and clustering quality metric (NMI) \cite{oh2017deep} with the same network architecture. Results are presented in Tab.~\ref{tab:KT3DMoSeg}.

Our vanilla approach achieved very competitive performance on all 22 sequences in KT3DMoSeg. In the `Augment' setting, our approach even outperforms the state-of-the-art multi-view clustering approaches (MVC) \cite{xu2018motion}. Of all benchmarked methods, only MVC has considered the multi-type fitting issue. 
Furthermore, we notice that our proposed MIMI metric is the best among all alternative losses considered. The unsupervised deep approaches lag behind by a large margin corroborating our earlier argument about the necessity to exploit labelled information for complex multi-type subspace clustering problem. It is obvious the deep approaches are very efficient in inference, costing only 1.85 seconds to process all sequences (from trajectory input to clustering output) while the best performing non-deep approach (MVC) costs 143.52 seconds. The only faster algorithm (GPCA) has a much worse performance.\\
Finally, we present qualitative comparisons in Fig.~\ref{fig:KT3DMoSeg}. The SubspaceNet surpasses our expectations in how it performs in `Seq009\_Clip01'. Here the independently moving car (the yellow group in the ground truth image) has a flow field that is consistent with the epipolar constraint associated with the background motion (due to them both translating in the same direction) \cite{xu2018motion}. Without resorting to reconstructing the depth of the car, it would be impossible to separate it from the background. However, criteria involving depth would be very unwieldy to specify analytically in the existing approaches. Here, without having any preconceived notion of the geometrical model, our network has learnt the requisite criteria to separate the independent motion.

\subsection{Non-Rigid Motion Segmentation}\label{sec:nonrigid}
We demonstrate the ability to learn non-rigid motion segmentation which is hard to be modeled by analytic geometric models. We train our model on the training set with 29 unique sequences and evaluate on the test set with 30 sequences following the rule established by \cite{ochs2014segmentation}. For our method, the number of motion is estimated via SOD\cite{Zhang2012} with candidate cluster range from 1 to 10. SSC\cite{Elhamifar2013}, ALC\cite{Rao2010}, Spectral Clustering (SC)\cite{ochs2014segmentation} and MultiCut\cite{magri2016multiple} are compared and the results are presented in Tab.~\ref{tab:FBMS}. We observe that our SubspaceNet, for both losses, is superior in performance compared with all three baseline methods.
We further present qualitative comparisons with \cite{ochs2014segmentation,magri2016multiple} in Fig.~\ref{fig:FBMS_Qualitative}. It is evident that the translational model employed in \cite{ochs2014segmentation} with spatial and color information \cite{magri2016multiple} detects the whole background but at the cost of over-segmenting the non-rigid foreground, e.g. the lion's head and the tractor's wheels. In contrast, our SubspaceNet detects the whole non-rigid foreground while keeping the background segmentation intact. Some objects are missed by all methods, e.g. the horse in stable of ``Farm01'', since it does not move significantly in the first 10 frames.


\begin{table*}[!htb]
        \begin{minipage}{0.66\textwidth}
            \centering
            \caption{\small{AdelaideRMF two-view motion segmentation classification error ($\%$). Inf. Time is inherited from \cite{tiwari2018dgsac} (- indicates unreported).}}
            \vspace{-0.3cm}
            \setlength\tabcolsep{2pt} 
	  \resizebox{1\linewidth}{!}{
   \begin{tabular}{lllllllllll}
    \toprule
          & \multicolumn{7}{c}{State-of-the-Arts}                 & \multicolumn{3}{c}{SubspaceNet} \\
\cmidrule(lr){2-8} \cmidrule(lr){9-11}          & J-Lnk\cite{toldo2008robust} & T-Lnk\cite{Magri2014} & RCMSA\cite{pham2014random} & RPA\cite{Magri2015}   & ILP\cite{magri2016multiple}   & DGSAC\cite{tiwari2018dgsac} & NMU\cite{tepper2017nonnegative}   & L2    & CE    & MIMI \\
    \midrule
    Average & 16.43 & 9.36  & 12.37 & \textbf{5.49}  & 6.04  & 6.95  & 5.72  & 6.13 &  6.92  & \textbf{5.17} \\
    Median & 14.29 & 7.80  & 9.87  & 4.57  & 4.27  &  4.98  &      \textbf{3.64} &  4.50   &  5.21   &  \textbf{3.00} \\
    Inf.Time & -     & 5.31  & -     & 967.2 & 145.9 & 114.72 & 499.6 & 1.4   & 1.4   & \textbf{1.4} \\
    \bottomrule
    \end{tabular}%
    }\label{tab:AdelaideRMF}
        \end{minipage}
        \hfill
        \begin{minipage}{0.34\textwidth}
        \centering
            \caption{\small{FBMS testset first 10 frames performance ($\%$).}}
            \vspace{-0.1cm}
   \setlength\tabcolsep{2pt} 
	  \resizebox{1\textwidth}{!}{
        \begin{tabular}{lrrrrrr}
    \toprule
    \multicolumn{1}{c}{\multirow{2}[4]{*}{Model}} & \multicolumn{1}{c}{\multirow{2}[4]{*}{SC\cite{ochs2014segmentation}}} & \multicolumn{1}{c}{\multirow{2}[4]{*}{SSC\cite{Elhamifar2013}}} & \multicolumn{1}{c}{\multirow{2}[4]{*}{ALC\cite{Rao2010}}} & \multicolumn{1}{c}{\multirow{2}[4]{*}{MC\cite{keuper2015motion}}} & \multicolumn{2}{c}{SubspaceNet} \\
\cmidrule{6-7}          &       &       &       &       & \multicolumn{1}{c}{L2} & \multicolumn{1}{c}{MIMI} \\
    \midrule
    Precision & 87.44 & 53.11 & \textbf{91.67} & 89.05 & 85.62 &  {85.70} \\
    Recall & 60.77 & 56.40  & 50.57 & 61.81 & {69.09} & \textbf{69.20} \\
    Fmeasure & 71.71 & 54.70  & 65.18 & 72.97 & {76.47} &   \textbf{76.57} \\
    \bottomrule
    \end{tabular}%
    }
  \label{tab:FBMS}%
        \end{minipage}
        \vspace{-0.5cm}
    \end{table*}

\subsection{Two-View Motion Segmentation}
We evaluate the motion segmentation task in the Adelaide RMF dataset \cite{wong2011dynamic}. We carry out a leave-one-out cross-validation. For comparability, we report the classification error rate (ErrorRate). The state-of-the-art models being compared include J-Linkage (J-Lnk)\cite{toldo2008robust}, T-Linkage (T-Lnk)\cite{Magri2014}, RPA \cite{Magri2015}, RCMSA \cite{pham2014random}, ILP-RansaCov (ILP)\cite{magri2016multiple}, DGSAC \cite{tiwari2018dgsac} and NMU\cite{tepper2017nonnegative}. The comparisons are presented in Tab.~\ref{tab:AdelaideRMF}. We observe that our SubspaceNet gives competitive results; in particular, our model  with MIMI loss gives a mean error of $5.17\%$. We note the performance is achieved by training on only a very small amount of data (18 sequences) and without any dataset-specific parameter tuning. We also notice that our SubspaceNet is very efficient at the inference stage (1.4 seconds) which is 30 times faster than the closest performing method (NMU costs 499.6 seconds). 

\subsection{Sampling Imbalance}

\begin{figure}[!htb]
\centering
\includegraphics[width=0.8\linewidth]{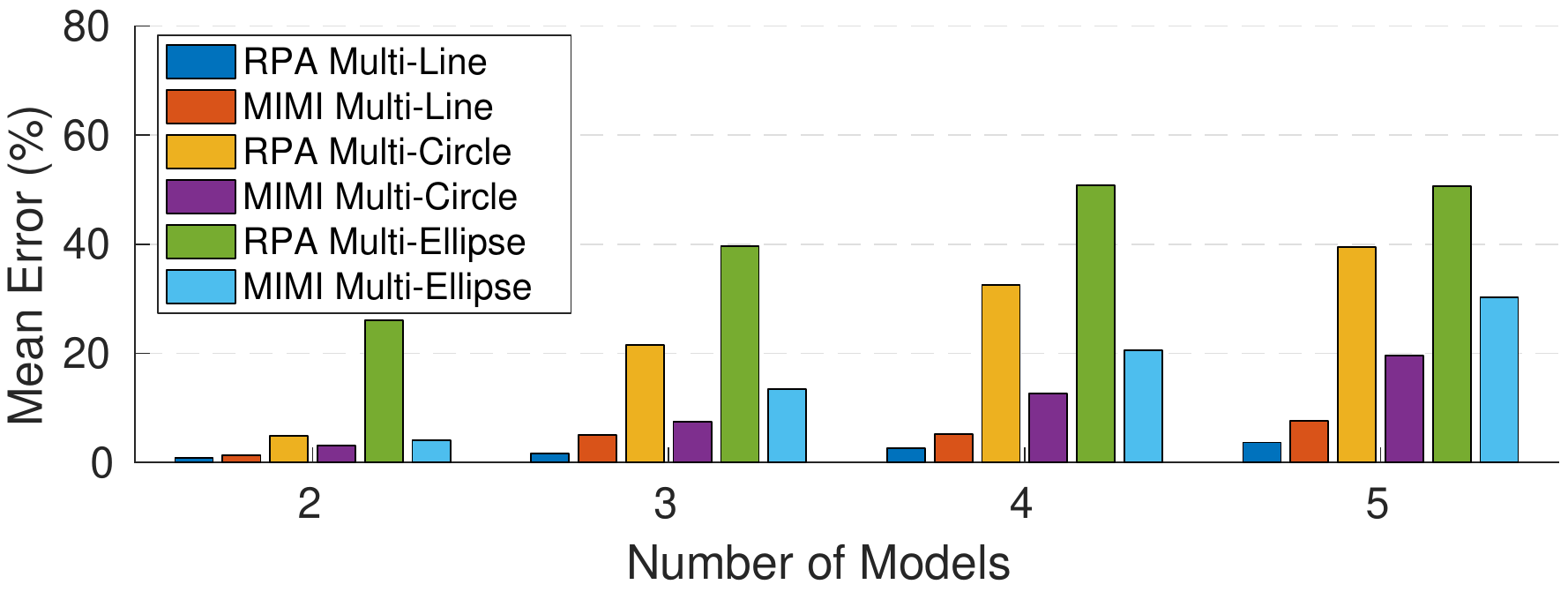}
\vspace{-0.3cm}
\caption{\small{Performance v.s. the number of models for synthetic multi-model fitting.}}\label{fig:SyntheticMultiModel}
\vspace{-0.5cm}
\end{figure}

In this section, we further demonstrate the ability of our network to robustly handle sampling imbalance, i.e. the inlier points represent a minority. We demonstrate via a synthetic single-type multi-model fitting problems. Specifically, we synthesize 8,000 training samples and 200 testing samples for each of the type, line, circle and ellipses, and compare with RPA\cite{Magri2015}. The results are presented in Fig.~\ref{fig:SyntheticMultiModel}. We conclude that, first, our multi-model network  performs comparably with RPA on multi-line segmentation task while outperforming RPA with large margin on the more challenging multi-circle and multi-ellipse tasks. The performance drops sharply from multi-line (blue) to multi-ellipse (green) fitting for RPA, with the drop getting more acute as the number of model increases. This suggests that the increasing size of the minimal support set (2 points for line, 3 points for circle and 5 points for ellipse) poses great challenge for the RANSAC-based approaches due to sampling imbalance. More precisely, in a noiseless $N$-model experiment, the chance of hitting the true model in a single sampling reduces from $(1/N)^2$ for straight line to $(1/N)^5$ for ellipse. It is evident that our multi-model network is less sensitive to the complexity of the model, as the drop in performance (purple and cyan bars) is less significant.

\subsection{Model Selection}
\label{sec:model_selection}

As can be seen from Fig.~\ref{fig:KT3DMoSeg}, the point distribution in the learned feature embedding is amenable for model selection. We evaluate the ability of both Second Order Difference (SOD)\cite{liu2013robust} and Silhouette Analysis (Silh.) \cite{rousseeuw1987silhouettes} to estimate the number of motions. We also compare with alternative subspace clustering approaches with built-in model selection, namely, LRR \cite{liu2013robust}, MSMC \cite{Dragon2012}, SSC\cite{Elhamifar2013}, GPCA\cite{vidal2005generalized}, ALC\cite{Rao2010} and additionally apply self-tuning spectral clustering(S.T.) \cite{zelnik2005self} to the affinity matrix obtained in MVC \cite{xu2018motion}. Among the above competitors, the model selection for GPCA and SSC are implemented with SOD. Performances are evaluated in terms of mean classification error (Mean Err) and correct rate (Correct), i.e. the percentage of samples/sequences with correctly estimated number of cluster (higher the better). Comparisons are presented in Tab.~\ref{tab:MdlSel}. Thanks to the deep feature learning, both SOD and Silh. applied to our method yield substantially better performance without the need to tune any parameter.


\begin{table}[htbp]
    \centering
\caption{\small{Comparison of model selection on KT3DMoSeg. Numbers are in $\%$.}}    \vspace{-0.3cm}
\setlength\tabcolsep{2pt} 
  \resizebox{0.99\linewidth}{!}{
    \begin{tabular}{lcccccccc}
    \toprule
    \multirow{2}[4]{*}{\textbf{Method}} & \multicolumn{2}{c}{\textbf{MIMI Loss}} & \multirow{2}[4]{*}{S.T.\cite{zelnik2005self}} & \multirow{2}[4]{*}{LRR\cite{liu2013robust}} & \multirow{2}[4]{*}{MSMC\cite{Dragon2012}} & \multirow{2}[4]{*}{ALC\cite{Rao2010}} & \multirow{2}[4]{*}{GPCA\cite{vidal2005generalized}} & \multirow{2}[4]{*}{SSC\cite{Elhamifar2013}} \\
\cmidrule{2-3}          & SOD\cite{liu2013robust}   & Silh.\cite{rousseeuw1987silhouettes} &       &       &       &       &       &  \\
    \midrule
    {Mean Err $\downarrow$} & 7.36  & \textbf{7.25} & 18.16 & 25.08 & 48.29 & 34.72 & 47.35 & 64.82 \\
    {Correct $\uparrow$} & \textbf{86.36} & 81.82 & 40.91 & 54.55 & 22.73 & 45.45 & 18.18 & 18.18 \\
    \bottomrule
    \end{tabular}%
    }
  \label{tab:MdlSel}%
  \vspace{-0.4cm}
\end{table}%

\subsection{Further Study}

\noindent\textbf{Feature Embedding}
We provide direct visualization of the learnt representations. We use T-SNE\cite{maaten2008visualizing} to project  both  the KT3DMoSeg raw feature points (of dimension ten for 5 frames) and network output embeddings to a 2-dimensional space. Three example sequences are presented in the last row of Fig.~\ref{fig:KT3DMoSeg}. We conclude from the figure that: (i) the original feature points are hard to be grouped by K-means correctly; and (ii) after our network embedding, feature points are more likely to be grouped according to the respective motions, regardless of the underlying types of motions.

\begin{figure}
\begin{center}
\subfloat{\includegraphics[width=0.33\linewidth]{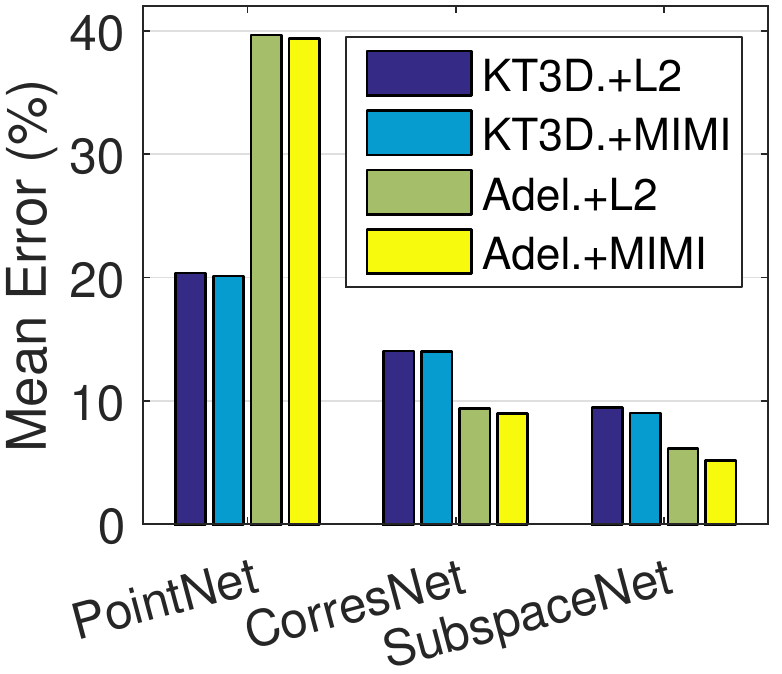}}
\subfloat{
\includegraphics[width=0.34\linewidth]{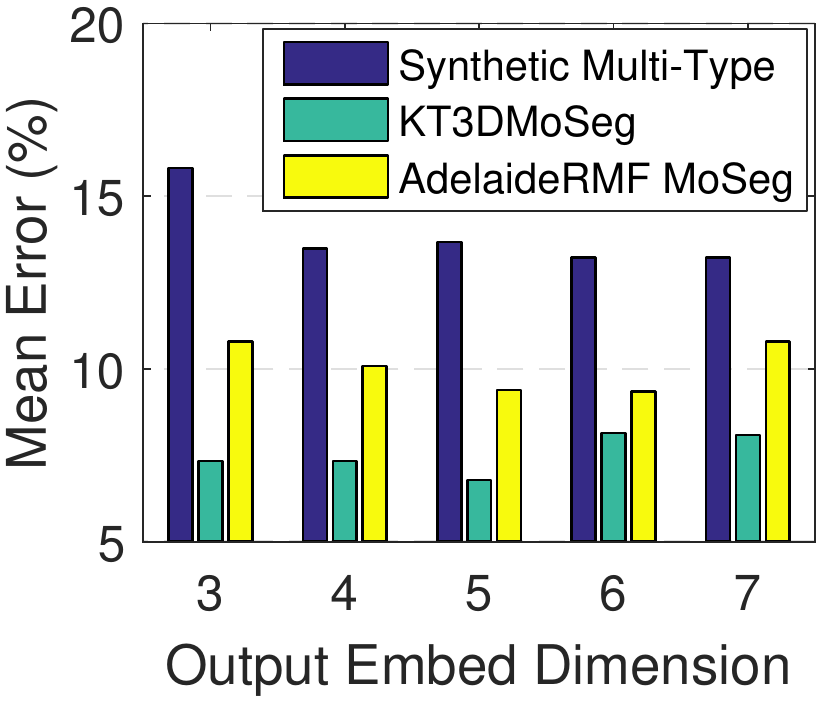}}
\subfloat{
\includegraphics[width=0.34\linewidth]{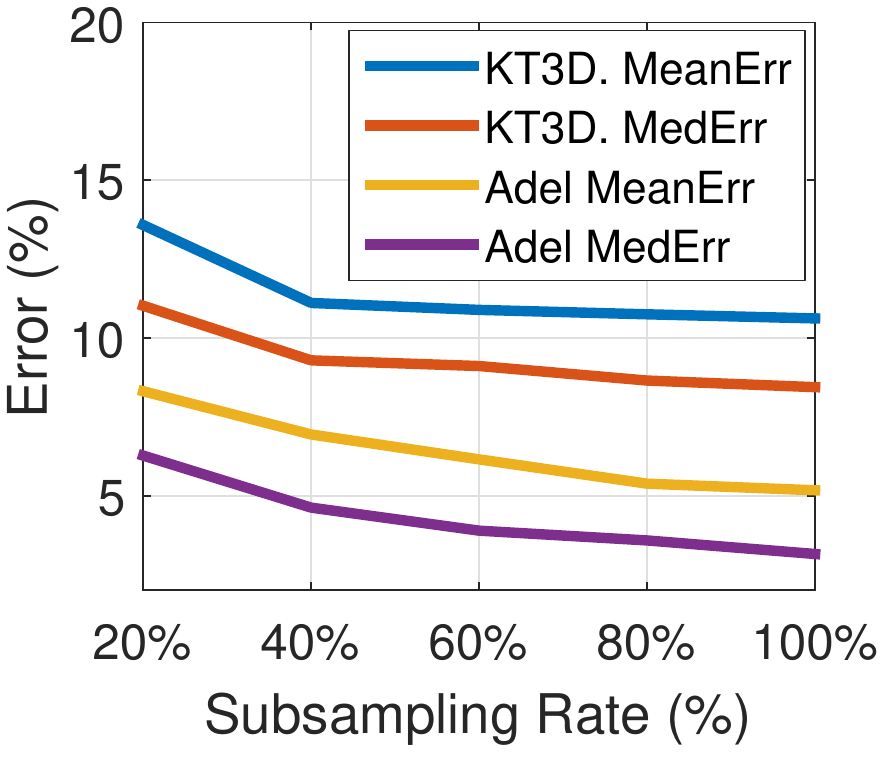}}\vspace{-0.3cm}
\caption{\small{(Left) Comaprison with alternative networks. (Middle) Performance v.s. the network output dimension. (Right) Weak supervision.}}\label{fig:Ablation}
\end{center}
\vspace{-0.6cm}
\end{figure}

\noindent\textbf{Dimension of Output Embedding}:
We investigate the impact of the dimension of the output embedding $\vect{z}$. 
We vary the size of the embedding dimension from 3 to 7 for three tasks and present the resulting error rates against the dimension in Fig~\ref{fig:Ablation} (middle). As can be seen, the errors are relatively stable w.r.t. the output
embedding dimension from 4 to 7 for all three tasks, with optimal
dimension between 5 to 6 coninciding with the maximal number of
clusters for each task (5 motions for KT3DMoSeg and
4 structures for Synthetic). Thus
the maximal number of clusters serves as a good heuristic for
the dimension of the network output embedding.


\noindent\textbf{Weak Supervision}: The SubspaceNet is trained on labelled data points which is often very costly to obtain compared with image category labels. In this section, we investigate the interaction between weaker supervision, i.e. fewer labelled data points and performance. In specific, we randomly subsample $20\%$ to $80\%$ labelled data points for each sequence in KT3DMoSeg and AdelaideRMF MoSeg and train the model with reduced labelled data while keeping the same evaluation protocol as normal. The results averaged over 10 trials are presented in Fig.~\ref{fig:Ablation} (right). We observe very stable error rate from $40\%$ subsample rate, suggesting the SubspaceNet is robust to fewer annotated data.


\noindent\textbf{Network Comparison}:
We compare SubspaceNet with alternative networks that are able to learn from sparse set of data. In particular, we compare with the correspondence network (CorresNet) \cite{yi2018learning} and PointNet \cite{Qi2017} on KT3DMoSeg(KT3D.) and AdelaideRMF MoSeg(Adel.), both of which are experimented with L2 Loss and our MIMI loss. The results are presented in Fig.~\ref{fig:Ablation}(left). We observe a significant performance gap between our SubspaceNet and the two alternatives. The proposed MIMI loss is also effective with alternative networks.

\section{Conclusion}

In this work, we investigate training a deep neural network for general multi-type subspace clustering. We formulate the problem as learning non-linear feature embeddings that maximize the distance between points of different clusters and minimize the variance within clusters. For inference, the output features are fed into a K-means to obtain the grouping. Model selection is easily achieved by just analyzing the K-means residual in a parameter free manner. Experiments are carried out on both synthetic and real motion segmentation tasks. Comparison with state-of-the-art approaches proves that our network can better deal with multiple types of models simultaneously.  
 Our method is also less sensitive to sampling imbalance brought about by the increasing number of models, and it is highly efficient at inference stage. As future works, one could consider including additional texture and color information and adopting sliding window technique to handle arbitrary long sequences.


{\small
\bibliographystyle{ieee}
\bibliography{egbib}
}

\newpage

\noindent\huge{\textbf{Supplementary}}

\normalsize
\section{Network Design}

In this section, we make more ablation study of the proposed SubspaceNet. In particular, we are concerned with the depth of the network, the necessity of the L2 normalization layer.

\subsection{Network Depth}
\begin{figure}[H]
\centering
\includegraphics[width=0.98\linewidth]{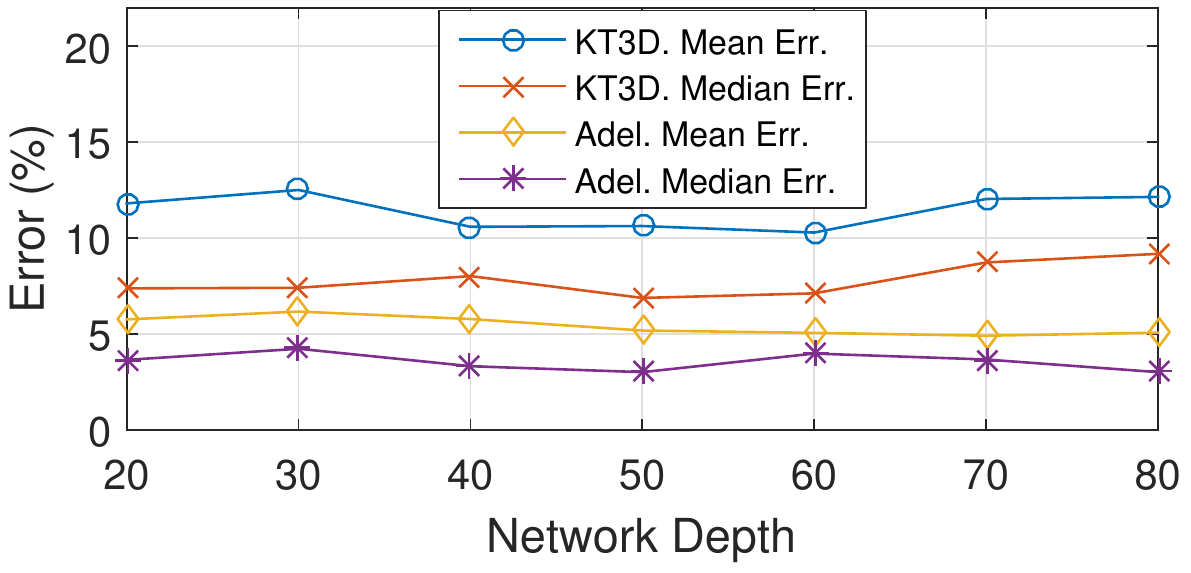}
\caption{The network depth v.s. the performance.}\label{fig:NetworkDepth}
\end{figure}

We evaluate the impact of the depth of SubspaceNet. The depth is varied from 20 to 80 with step of 10 and both the mean error and median error on KT3DMoSeg(KT3D.) and AdelaideRMF MoSeg (Adel.) are reported in Fig.~\ref{fig:NetworkDepth}. We observe relatively stable performance w.r.t the depth of network thanks to the ResNet structure. In particular, the optimal range is between 40 to 60.

\subsection{L2 Normalization Layer}
We introduced a L2 normalization layer at the output of SubspaceNet. This layer normalizes the scales of all feature embeddings so that all feature points lie on a unit sphere, thereby benefitting the metric learning procedure. We specifically evaluate the necessity of this layer by comparing the results on motion segmentation with and without the L2 normalization layer. As can be seen from Tab.~\ref{tab:CompareL2Norm},  the performance is consistently better with the L2norm layer for both the KT3DMoSeg and AdelaideRMF MoSeg datasets, suggesting that the L2norm layer is beneficial for learning better feature embeddings.

\begin{table}[htbp]
  \centering
  \caption{Comparison of with (w) L2norm layer and without (w/o) L2norm layer. The numbers are in $\%$.}
    \begin{tabular}{lrrrr}
    \toprule
          & \multicolumn{2}{c}{w L2norm} & \multicolumn{2}{c}{w/o L2norm} \\
\cmidrule{2-5}    Dataset & \multicolumn{1}{l}{Mean} & \multicolumn{1}{l}{Med.} & \multicolumn{1}{l}{Mean} & \multicolumn{1}{l}{Med.} \\
    \midrule
    KT3DMoSeg & 10.62 & 8.44  & 11.56 & 8.78 \\
    AdelaideMoSeg & 5.17  & 3.00  & 6.68  & 3.81 \\
    \bottomrule
    \end{tabular}%
  \label{tab:CompareL2Norm}%
\end{table}%

\subsection{Tensorflow Implementation}

\begin{figure*}
\centering
\caption{SubspaceNet code.}\label{fig:SubsapceNet}
\includegraphics[width=1.3\linewidth]{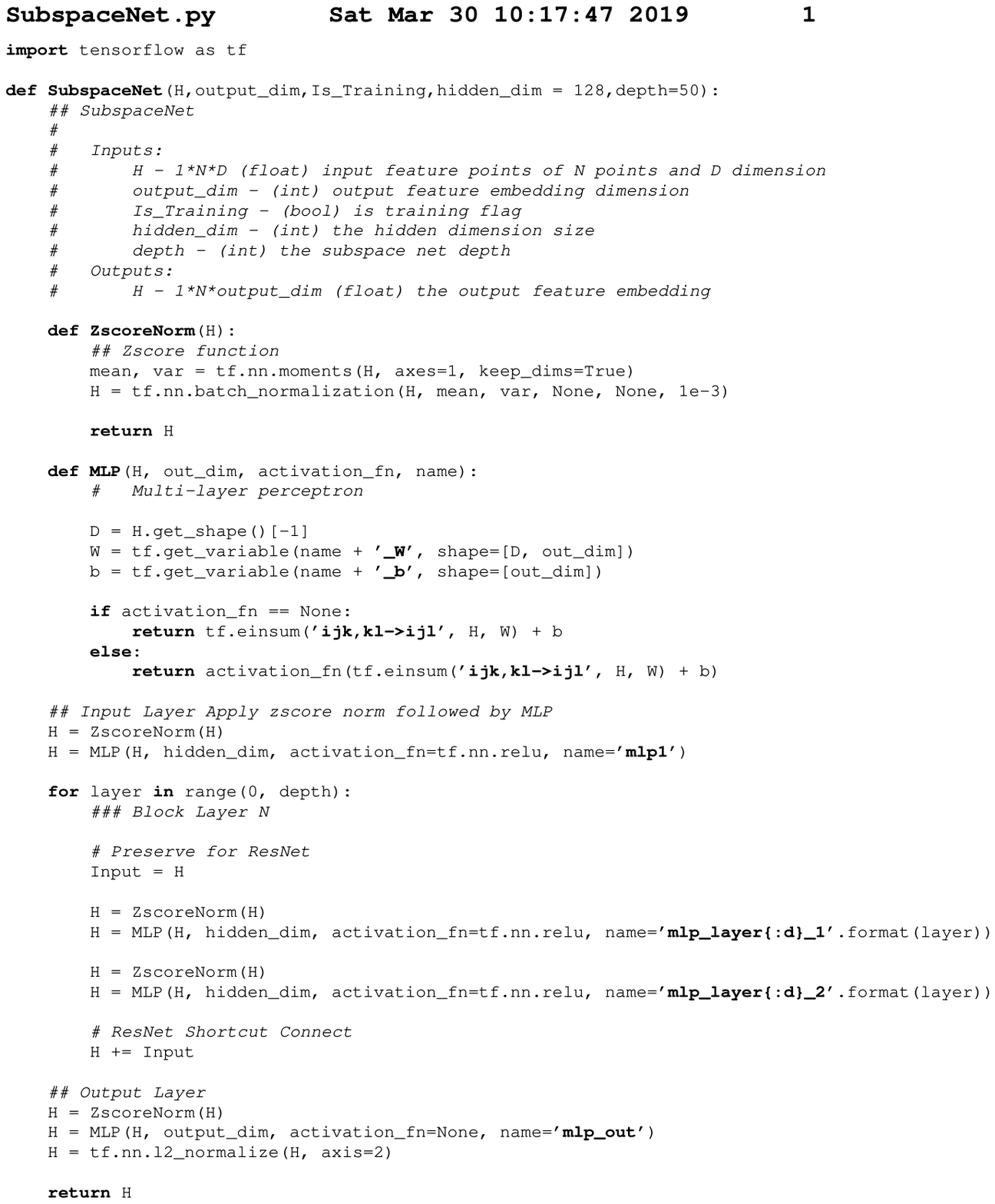}
\end{figure*}

We further append the tensorflow implementation of the proposed SubspaceNet in Fig.~\ref{fig:SubsapceNet}.

\section{Training}
\begin{figure}[!htb]
\centering
\includegraphics[width=1\linewidth]{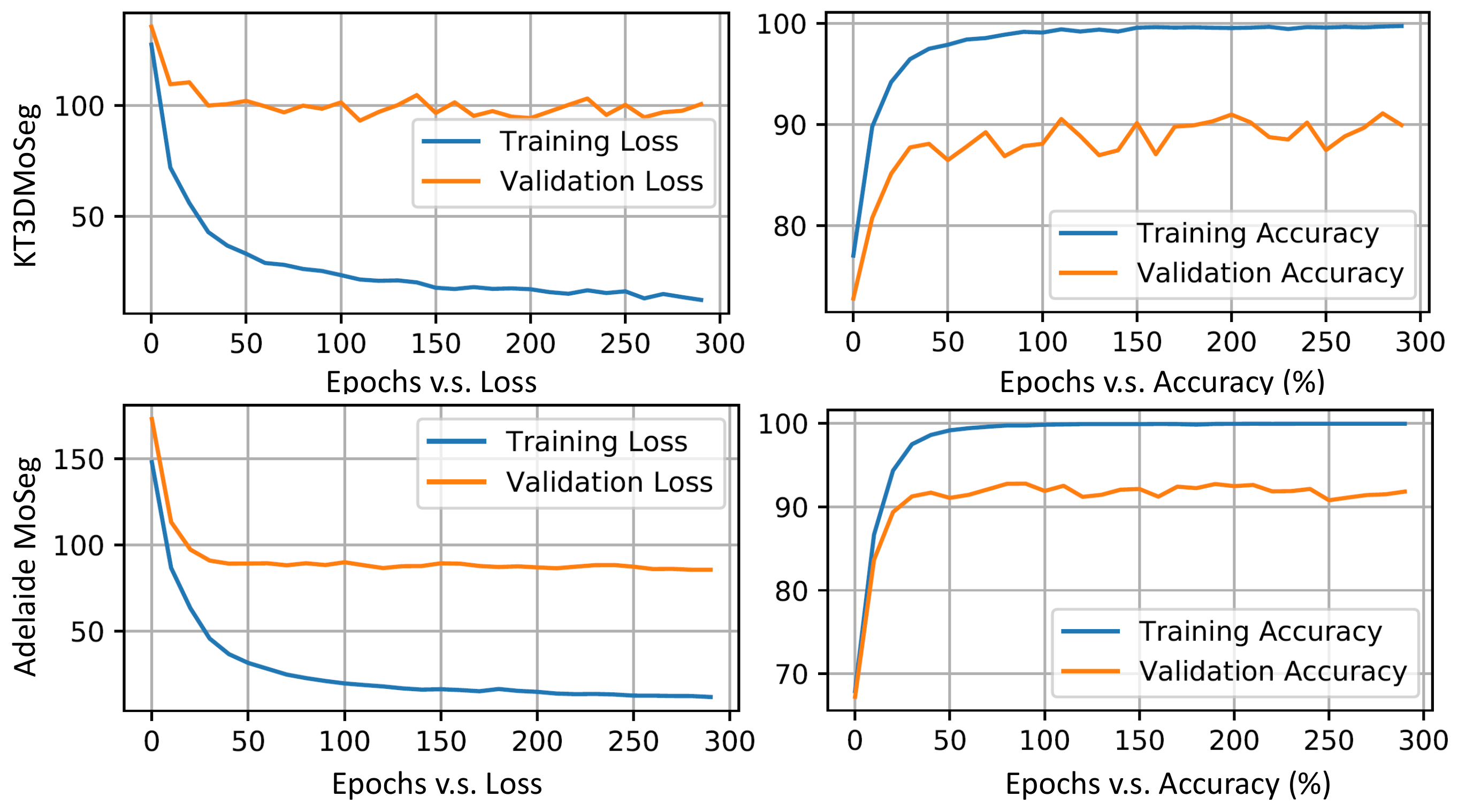}
\caption{Training procedure: epochs v.s. loss and accuracy.}\label{fig:TrainingProc}
\end{figure}

Due to the limited size of existing subspace clustering datasets, exhibiting a low diversity of motion-scene types for motion segmentation, one might suspect the risk of overfitting. In this section, we investigate this issue by visualizing both the training/validation loss and errors. The results on both KT3DMoSeg  and AdelaideRMF MoSeg are shown in Fig.~\ref{fig:TrainingProc}. We observe both training and validation loss converging after 100 epochs as does the prediction accuracy.

\begin{figure*}[h]
\begin{equation}
\resizebox{0.92\hsize}{!}
{$
\begin{split}
&\nabla_\Theta L(\Theta) = -\sum\limits_{i\in\mathcal{C}_m}\frac{\frac{1}{|\mathcal{C}_m|^2}\left(2\vect{z}_i + \sum\limits_{j\in\mathcal{C}_m,j\neq i} \vect{z}_j\right) - \frac{1}{|\mathcal{C}_m||\mathcal{C}_n|}\sum\limits_{j\in\mathcal{C}_n}\vect{z}_j}{||\bm{\mu}_m-\bm{\mu}_n||_2^2}\nabla_\Theta f(\matr{x}_i;\Theta) -\sum\limits_{j\in\mathcal{C}_n}\frac{\frac{1}{|\mathcal{C}_n|^2}\left(2\vect{z}_j + \sum\limits_{i\in\mathcal{C}_m,i\neq j} \vect{z}_i\right) - \frac{1}{|\mathcal{C}_n||\mathcal{C}_m|}\sum\limits_{k\in\mathcal{C}_m}\vect{z}_k}{||\bm{\mu}_m-\bm{\mu}_n||_2^2}\nabla_\Theta f(\matr{x}_j;\Theta)\\
&+\alpha\sum\limits_{k\in \mathcal{C}_l} \Big(2\vect{z}_k - \frac{1}{|\mathcal{C}_l|}\left(2\vect{z}_k + \sum\limits_{j\in\mathcal{C}_l,j\neq i}\vect{z}_j\right) + \frac{1}{|\mathcal{C}_l|^2}\left(2\vect{z}_k + 2\sum\limits_{j\in\mathcal{C}_l,j\neq i}\vect{z}_j\right)\Big)\nabla_\Theta f(\matr{x}_k;\Theta)
\end{split}
$}
\end{equation}\vspace{-0.5cm}
\end{figure*}\label{eq:MIMI_Gradient}

\section{Loss Design}

In this section, we present more analysis and derivations of the proposed MaxInterMinIntra (MIMI) loss. Specifically, we first provide the gradient of the MIMI loss and then analyze the two components of the MIMI loss. Finally, we publish the tensorflow implementation of the proposed MaxInterMinIntra code.

\subsection{Gradient of MIMI Loss}
We explicitly provide the gradient of the proposed MaxInterMinIntra loss with respect to the output of SubspaceNet in Eq~.(\ref{eq:MIMI_Gradient}). The MIMI loss can be efficiently optimized via backpropagation.


\subsection{MIMI Loss Components}
\begin{figure}[H]
\includegraphics[width=0.9\linewidth]{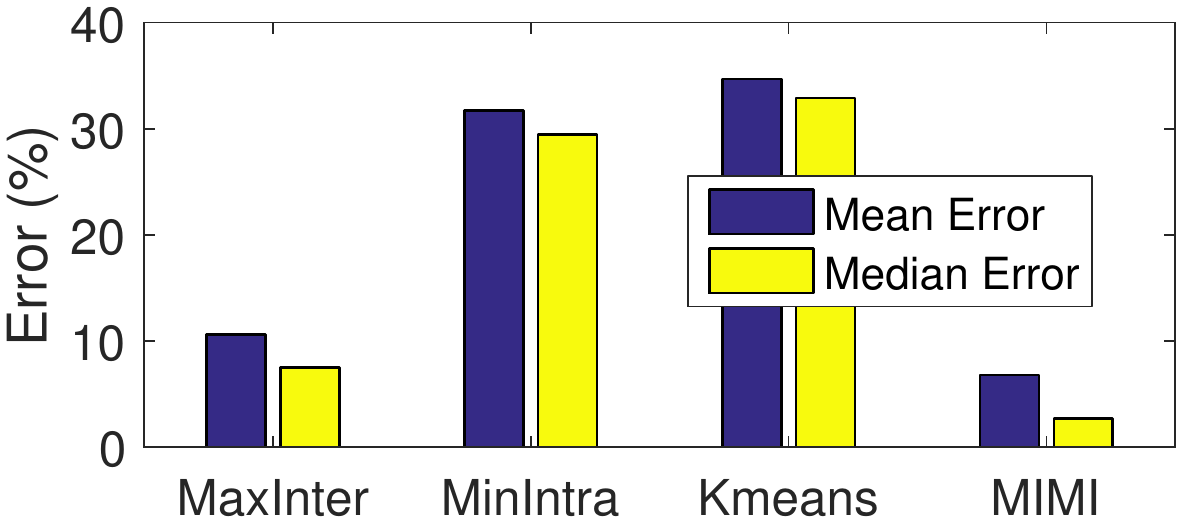}
\caption{Comparison of different variants of MIMI loss.}\label{fig:MIMI}
\end{figure}

Here we investigate the necessity of both maximizing inter cluster distance and minimizing intra cluster variance. Specifically, we compare the following variants. (i) \textbf{MaxInter}: only maximizing the inter cluster distance is considered, equivalent to the first term in Eq~(\ref{eq:MaxInterMinIntra}). (ii) \textbf{MinIntra}: only minimizing the intra cluster variance is considered, the second term in Eq~(\ref{eq:MaxInterMinIntra}). (iii) \textbf{K-means loss}: we further note the k-means loss \cite{yang2017towards} proposed for unsupervised deep clustering shares the same objective with \textbf{MinIntra}. We therefore adapt the k-means loss to supervised learning with fixed point-to-cluster assignment during training. We compare the three variants with our final MIMI loss on KT3DMoSeg and present the results in Fig.~\ref{fig:MIMI}. The MIMI loss is consistently better (lower error) than all three variants. In particular, the \textbf{MinIntra} and \textbf{K-means loss} produce large errors. This indicates that pushing points of different clusters away is vital to feature embedding for clustering.

\begin{equation}\label{eq:MaxInterMinIntra}
L(\Theta) = -\log \min\limits_{m,n} ||\bm{\mu}_m-\bm{\mu}_n||_2^2 + \log\max_{l} s_l
\end{equation}

\subsection{Tensorflow Implementation}

\begin{figure*}
\centering
\caption{MaxInterMinIntra code.}\label{fig:MaxInterMinIntraCode}
\includegraphics[width=1.1\linewidth]{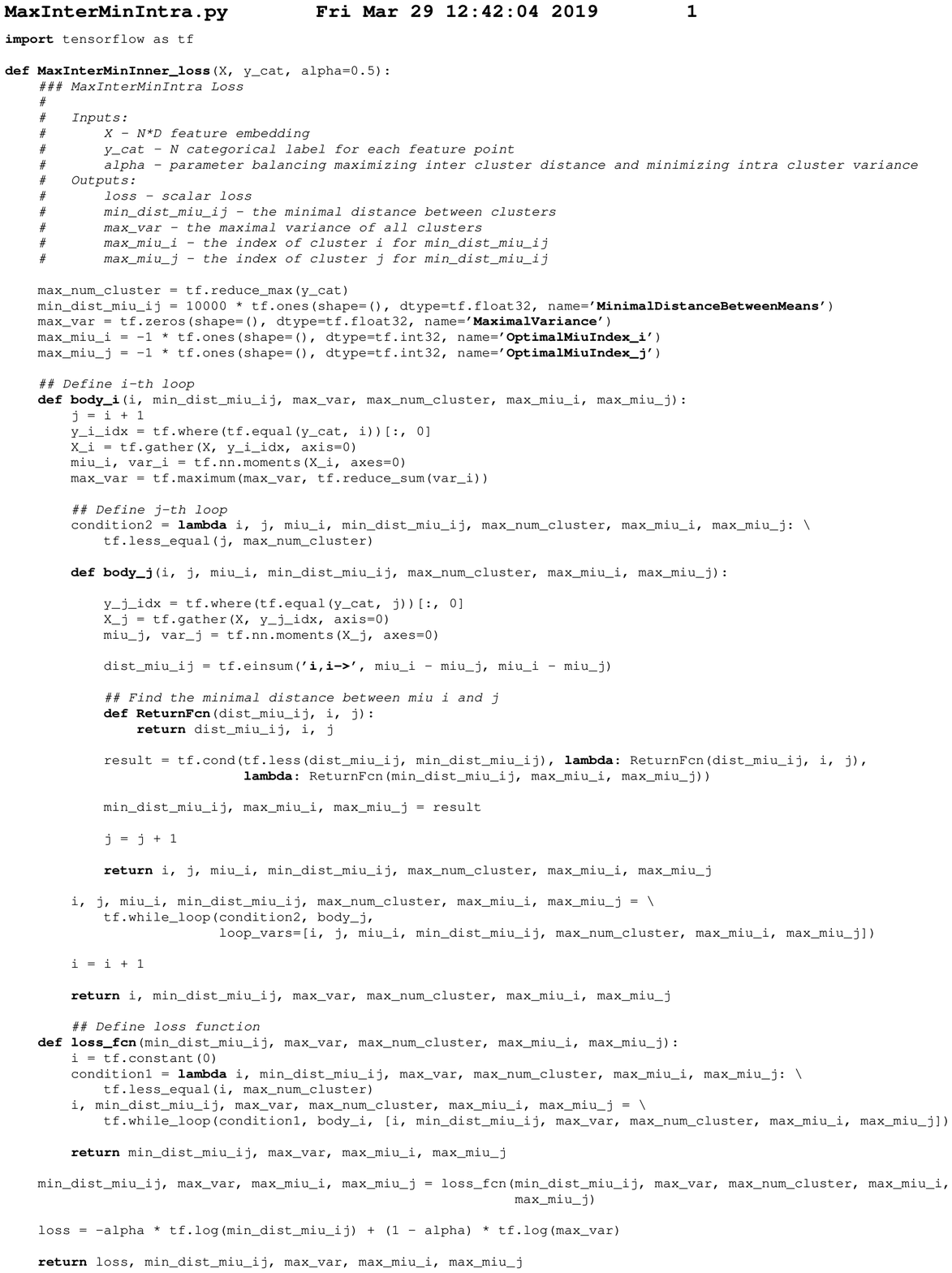}
\end{figure*}

The tensorflow implementation of the proposed MaxInterMinIntra loss is shown in Fig.~\ref{fig:MaxInterMinIntraCode}.

\section{Additional Results}
\begin{figure*}[!h]
\centering
\includegraphics[width=1.05\linewidth]{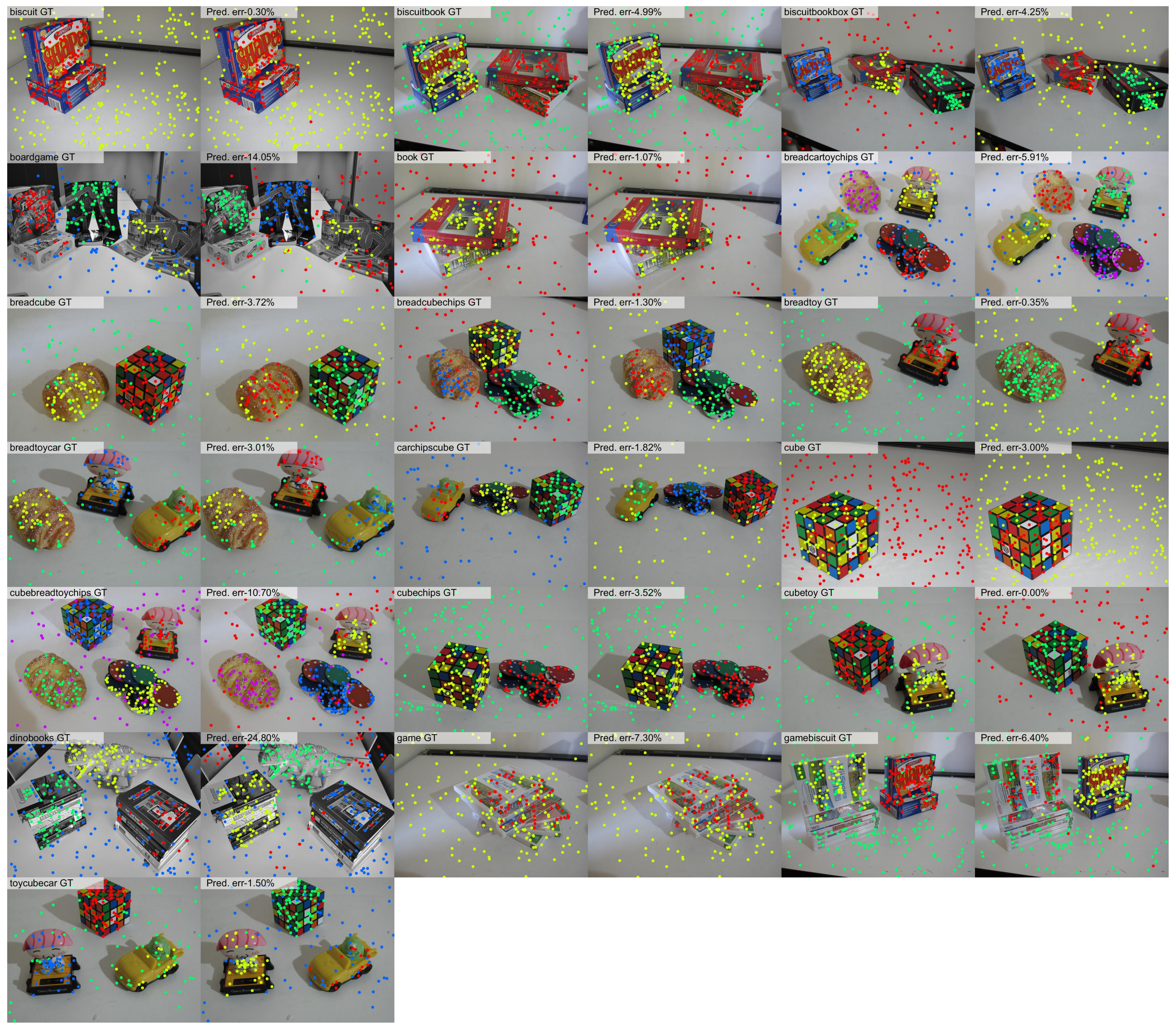}
\caption{Qualitative evaluation of AdelaideRMF MoSeg dataset.}\label{fig:AdelaideMoSeg}
\end{figure*}

We present in this section additional qualitative results on the datasets evaluated in the manuscript. In particular, we demonstrate the motion segmentation results on AdelaideRMF MoSeg dataset in Fig.~\ref{fig:AdelaideMoSeg}. For each sequence, we visualize both the ground-truth as `XXX GT' and `Pred. err-Y.YY\%' where `XXX' is the sequence name and `Y.YY' is the error rate. We also report the per sequence performance and compare with multiple state-of-the-art methods in Tab.~\ref{tab:AdelaideMoSeg}. We observe very competitive performance on most sequences of AdelaideRMF MoSeg dataset with only one sequence (dinobooks) missing one book. It is also noteworthy that our approach does generalize to new shapes well. For example, the boardgames in the  sequence `boardgame' (middle\& right), the dinosaur in the sequence `dinobooks' (middle) and the stacked games in the sequence `game' all have never appeared in any other sequences. For the shapes that appear in multiple sequences, we notice the extracted feature points are quite different from sequence to sequence. For example, the toys, cubes and bread in all different sequences are very different in term of feature points thus overfitting to the shape is not allowed.

\begin{table*}[htbp]
  \centering
  \caption{Per sequence performance on AdelaideRMF MoSeg. The accuracy ($\%$) is reported for each method and the numbers are inherited from \cite{tiwari2018dgsac}. k is the number of ground-truth foreground. }
  \footnotesize
   \setlength\tabcolsep{1pt} 
	  \resizebox{1.01\linewidth}{!}{
    \begin{tabular}{l|c|c|c|c|c|c|c|c|c|c|c|c|c|c|c|c|c|c|c|c|c}
    \toprule
    \multicolumn{1}{r}{} & \multicolumn{1}{l}{\begin{sideways}'biscuit'\end{sideways}} & \multicolumn{1}{l}{\begin{sideways}'biscuitbook'\end{sideways}} & \multicolumn{1}{l}{\begin{sideways}'biscuitbookbox'\end{sideways}} & \multicolumn{1}{l}{\begin{sideways}'boardgame'\end{sideways}} & \multicolumn{1}{l}{\begin{sideways}'book'\end{sideways}} & \multicolumn{1}{l}{\begin{sideways}'breadcartoychips'\end{sideways}} & \multicolumn{1}{l}{\begin{sideways}'breadcube'\end{sideways}} & \multicolumn{1}{l}{\begin{sideways}'breadcubechips'\end{sideways}} & \multicolumn{1}{l}{\begin{sideways}'breadtoy'\end{sideways}} & \multicolumn{1}{l}{\begin{sideways}'breadtoycar'\end{sideways}} & \multicolumn{1}{l}{\begin{sideways}'carchipscube'\end{sideways}} & \multicolumn{1}{l}{\begin{sideways}'cube'\end{sideways}} & \multicolumn{1}{l}{\begin{sideways}'cubebreadtoychips'\end{sideways}} & \multicolumn{1}{l}{\begin{sideways}'cubechips'\end{sideways}} & \multicolumn{1}{l}{\begin{sideways}'cubetoy'\end{sideways}} & \multicolumn{1}{l}{\begin{sideways}'dinobooks'\end{sideways}} & \multicolumn{1}{l}{\begin{sideways}'game'\end{sideways}} & \multicolumn{1}{l}{\begin{sideways}'gamebiscuit'\end{sideways}} & \multicolumn{1}{l}{\begin{sideways}'toycubecar'\end{sideways}} & \multicolumn{1}{l}{Mean} & \multicolumn{1}{l}{Median} \\
    \midrule
    Outlier (\%) & 57.16 & 47.51 & 37.21 & 42.48 & 21.48 & 35.2  & 32.19 & 35.22 & 37.41 & 34.15 & 36.59 & 69.49 & 28.03 & 51.62 & \multicolumn{1}{c}{41.42} & 44.54 & 73.48 & 51.54 & 36.36 &       &  \\
    k     & 1     & 2     & 3     & 3     & 1     & 4     & 2     & 3     & 2     & 3     & 3     & 1     & 4     & 2     & \multicolumn{1}{c}{2} & 3     & 1     & 2     & 3     &       &  \\
    Tlnk  & 83.09 & 97.77 & 88.80 & 83.73 & 82.57 & 80.51 & 85.62 & 82.00 & 96.81 & 84.70 & 88.00 & 46.29 & 80.18 & 95.14 & 78.80 & 78.56 & 77.60 & 70.61 & 70.70 & 81.65 & 82.57 \\
    RCM   & 95.15 & 92.52 & 83.71 & 78.46 & 94.01 & 78.82 & 87.27 & 83.17 & 78.37 & 83.07 & 78.85 & 87.98 & 81.62 & 90.32 & 89.64 & 72.28 & 90.77 & 85.40 & 83.45 & 84.99 & 83.71 \\
    RPA   & 98.36 & 96.42 & 95.83 & 87.53 & 97.54 & 91.73 & 95.95 & 95.57 & 97.15 & 92.17 & 94.30 & 97.15 & 93.21 & 96.48 & 96.31 & 84.78 & 95.97 & 96.95 & 91.70 & 94.47 & 95.95 \\
    DPA   & 82.12 & 97.24 & 95.14 & 83.69 & 90.16 & 91.56 & 94.09 & 94.61 & 90.59 & 88.67 & 86.30 & 96.89 & 87.28 & 92.92 & 93.61 & 84.17 & 97.47 & 90.95 & 85.65 & 90.64 & 90.95 \\
    Cov   & 98.35 & 97.58 & 93.99 & 77.82 & 97.16 & 87.34 & 95.92 & 88.59 & 82.37 & 89.18 & 88.72 & 97.12 & 90.68 & 93.59 & 95.50 & 68.66 & 92.39 & 95.52 & 82.07 & 90.14 & 92.39 \\
    NMU   & 97.58 & 98.83 & 98.07 & 82.80 & 100.00 & 94.94 & 97.11 & 97.39 & 97.92 & 92.17 & 97.58 & 98.01 & 87.16 & 98.59 & 97.99 & 84.44 & 98.71 & 92.07 & 91.50 & 94.89 & 97.58 \\
    DGSAC & 98.18 & 98.58 & 97.58 & 82.59 & 99.03 & 87.88 & 97.73 & 93.04 & 90.74 & 89.49 & 85.52 & 96.76 & 88.57 & 97.38 & 97.32 & 83.92 & 95.02 & 98.19 & 90.35 & 93.05 & 95.02 \\
    Ours  & 99.70 & 95.01 & 95.75 & 85.95 & 98.93 & 94.09 & 96.28 & 98.70 & 99.65 & 96.99 & 98.18 & 97.00 & 89.30 & 96.48 & 100.00 & 75.20 & 92.70 & 93.60 & 98.50 & {94.83} & 97.00 \\
    \bottomrule
    \end{tabular}%
  \label{tab:AdelaideMoSeg}%
  }
\end{table*}%

\newpage
\newpage
\newpage

\end{document}